%% file: main.tex
\documentclass[sn-mathphys-num]{sn-jnl}% Math and Physical Sciences Numbered Reference Style 
\usepackage{graphicx}%
\usepackage{multirow}%
\usepackage{amsmath,amssymb,amsfonts}%
\usepackage{amsthm}%
\usepackage{mathrsfs}%
\usepackage[title]{appendix}%
\usepackage{xcolor}%
\usepackage{textcomp}%
\usepackage{manyfoot}%
\usepackage{booktabs}%
\usepackage{algorithm}%
\usepackage{algorithmicx}%
\usepackage{algpseudocode}%
\usepackage{listings}%
\usepackage{colortbl}
\usepackage{amsmath}
\usepackage{tikz} 
\usepackage{comment}
\usepackage{svg}
\svgsetup{inkscapelatex=false}

\usepackage{wrapfig}
\usepackage{cleveref}

\usepackage{subcaption}
\usepackage{collcell}
\usepackage{colortbl}
\usepackage{highlight}
\usepackage{amssymb}% http://ctan.org/pkg/amssymb
\usepackage{pifont}% http://ctan.org/pkg/pifont
\newcommand{\cmark}{\ding{51}}%
\newcommand{\xmark}{\ding{55}}%
\newcommand{\g}[1]{\gradientcelld{#1}{-1}{0}{1}{red}{white}{green}{70}}
\newcommand{\inv}[1]{}
\definecolor{customgray}{gray}{0.7} % 0 is black, 1 is white -> For tables
\definecolor{backcolour}{rgb}{0.95,0.95,0.92}

\newcommand{\rev}[1]{\textcolor{red}{#1}}
\lstnewenvironment{prompt}{%
  \thicklines
  \lstset{
    backgroundcolor=\color{backcolour},
    rulecolor=\color{backcolour!40!black},
    frameround=tttt,
    frame=single,
    basicstyle=\ttfamily\scriptsize,
    basewidth=0.50em,
    keywordstyle=\bfseries,
    xleftmargin=.02\linewidth,
    xrightmargin=.02\linewidth,
    escapeinside={<@}{@>},
    numbers=none
  }}{}

\begin{document}

\title[Linearly-Interpretable Concept Embedding Models for Text Analysis]{Linearly-Interpretable Concept Embedding Models for Text Analysis\footnote{Paper accepted at ECML-PKDD 2025 (Journal Track)}}


\author*[1]{\fnm{Francesco} \sur{De Santis}}\email{francesco.desantis@polito.it}
\equalcont{These authors contributed equally to this work.}

\author[2]{\fnm{Philippe} \sur{Bich}}\email{philippe.bich@polito.it}
\equalcont{These authors contributed equally to this work.}

\author[1,3]{\fnm{Gabriele} \sur{Ciravegna}}\email{gabriele.ciravegna@polito.it}
\equalcont{These authors contributed equally to this work.}

\author[4]{\fnm{Pietro} \sur{Barbiero}}\email{pietro.barbiero@usi.ch}

\author[1]{\fnm{Danilo} \sur{Giordano}}\email{danilo.giordano@polito.it}

\author[1]{\fnm{Tania} \sur{Cerquitelli}}\email{tania.cerquitelli@polito.it}

\affil[1]{ \orgdiv{Dipartimento di Automatica e Informatica}, \orgname{Politecnico di Torino}, \orgaddress{\street{Corso Duca degli Abruzzi}, \city{Torino}, \postcode{ 10129}, \country{Italy}}}

\affil[2]{\orgdiv{Dipartimento di Elettronica e Telecomunicazioni}, \orgname{Politecnico di Torino}, \orgaddress{\street{Corso Duca degli Abruzzi}, \city{Torino}, \postcode{ 10129}, \country{Italy}}}

\affil[3]{\orgname{CENTAI Institute}, \orgaddress{\street{ Corso Inghilterra, 3}, \city{Torino}, \postcode{10138}, \country{Italy}}}

\affil[4]{\orgdiv{Faculty of Informatics}, \orgname{Universita' della Svizzera Italiana}, \orgaddress{\street{ Via Giuseppe Buffi}, \city{Lugano}, \postcode{6900}, \country{Switzerland}}}


\abstract{Despite their success, Large-Language Models (LLMs) still face criticism due to their lack of interpretability.
Traditional post-hoc interpretation methods, based on attention and gradient-based analysis, offer limited insights as they only approximate the model's decision-making processes 
and have been proved to be unreliable.
For this reason, Concept-Bottleneck Models (CBMs) have been lately proposed in the textual field to provide interpretable predictions based on human-understandable concepts. 
However, CBMs still exhibit several limitations due to their architectural constraints limiting their expressivity, to the absence of task-interpretability when employing non-linear task predictors and for requiring extensive annotations that are impractical for real-world text data. In this paper, we address these challenges by proposing a novel Linearly Interpretable Concept Embedding Model (LICEM) going beyond the current accuracy-interpretability trade-off. LICEMs classification accuracy is better than existing interpretable models and matches black-box ones. We show that the explanations provided by our models are more intervenable and causally consistent with respect to existing solutions. 
Finally, we show that LICEMs can be trained without requiring any concept supervision, as concepts can be automatically predicted when using an LLM backbone.}

\keywords{Concept-XAI, Text Analysis, Linear Concept Attribution}



\maketitle

\section{Introduction}
In recent years, Large-Language Models (LLMs) have revolutionized the way we approach text interpretation, generation, and classification~\citep{devlin2018bert, brown2020language}.
Despite their success, LLMs' reliability is insufficient, due to the occurrence of hallucinations~\citep{huang2023survey} and the inconsistency of self-provided explanations that often do not reflect the actual decision-making process~\citep{ye2022unreliability, madsen2024self}. Traditional explainability methods rely mainly on the post-hoc analysis of the attention mechanisms~\citep{jain2019attention, wiegreffe2019attention} and the output gradients~\citep{chefer2021transformer}, both of which have shown limited interpretability as they are often unreliable~\citep{adebayo2018sanity, taimeskhanov2024cam} and only show \textit{where} the model looks, but not \textit{what} it sees in a given input~\citep{rudin2019stop, poeta2023concept}.

%These approaches only offer insight about which tokens mostly influenced the model prediction, but whether they explain the model's decision-making process is disputed~\citep{rudin2019stop, adebayo2018sanity}, and  more importantly, do not allow human experts to intervene upon the model to avoid hallucinations.
% they cannot control the generation of hallucinations.  

For this reason, Concept-Bottleneck Models (CBMs)~\citep{koh2020concept} have been recently proposed in the textual field to improve the interpretability of LLM predictions~\citep{tan2024interpreting, tan2024sparsity}.  In CBMs, an intermediate layer outputs a set of human-understandable symbols, commonly referred to as concepts, before providing the final classification. %\authorcomment{FDS}{blue}{
While CBMs utilize a black-box module to predict concepts, they enhance the interpretability of end-to-end (E2E) deep neural networks by providing a transparent intermediate representation that allows users to interact with the model~\citep{koh2020concept}. With CBMs, users can check and modify the predicted concepts to extract counterfactual predictions.
%in such a way they can interact with the model. %and gain insights into the relationship between concepts and labels.
%}
%Furthermore, CBMs allow by-design counterfactual predictions based on human-driven modifications of the predicted concepts (i.e., concept interventions). 
However, CBMs still present several limitations:
i)  the concept bottleneck architecture prevents high classification accuracy, especially in real-world text scenarios where complete concept representations~\citep{yeh2020completeness} are difficult to obtain; 
ii) when CBMs employ non-linear task predictors or provide predictions on top of concept embeddings~\cite{zarlenga2022concept}, they are not \textit{task-interpretable}, i.e., the decision process from the concepts to the final classification is non-interpretable;
iii) concept annotation in CBMs is expensive, and existing generative concept annotation approaches~\citep{tan2024interpreting} require the use of multiple modules.
%As shown in Figure~\ref{fig:viz-abstract} (left), concepts such as `Effectiveness' or `Side Effects' provide a very comprehensible means to interpret the model's decision process. In contrast, understanding the decision process by examining the importance of single tokens like `rid' or `worse' can be challenging. % without considering all the relations with the other tokens.  %which can be challenging to interpret without considering all the relations with the other tokens. %, particularly in relation to other tokens (e.g., `things'). 

\begin{figure}
        \centering
        % \includesvg[width=.4\textwidth]{LICEM.svg}
        \includegraphics[trim = 50 270 220 150, clip, width=.4\linewidth]{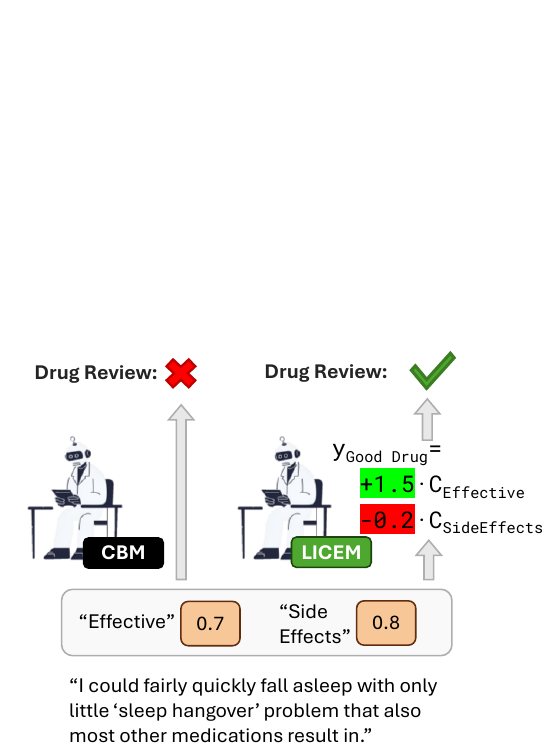}
        \hfill
        \includegraphics[trim = 0 -20 0 0,clip, width=.59\linewidth]{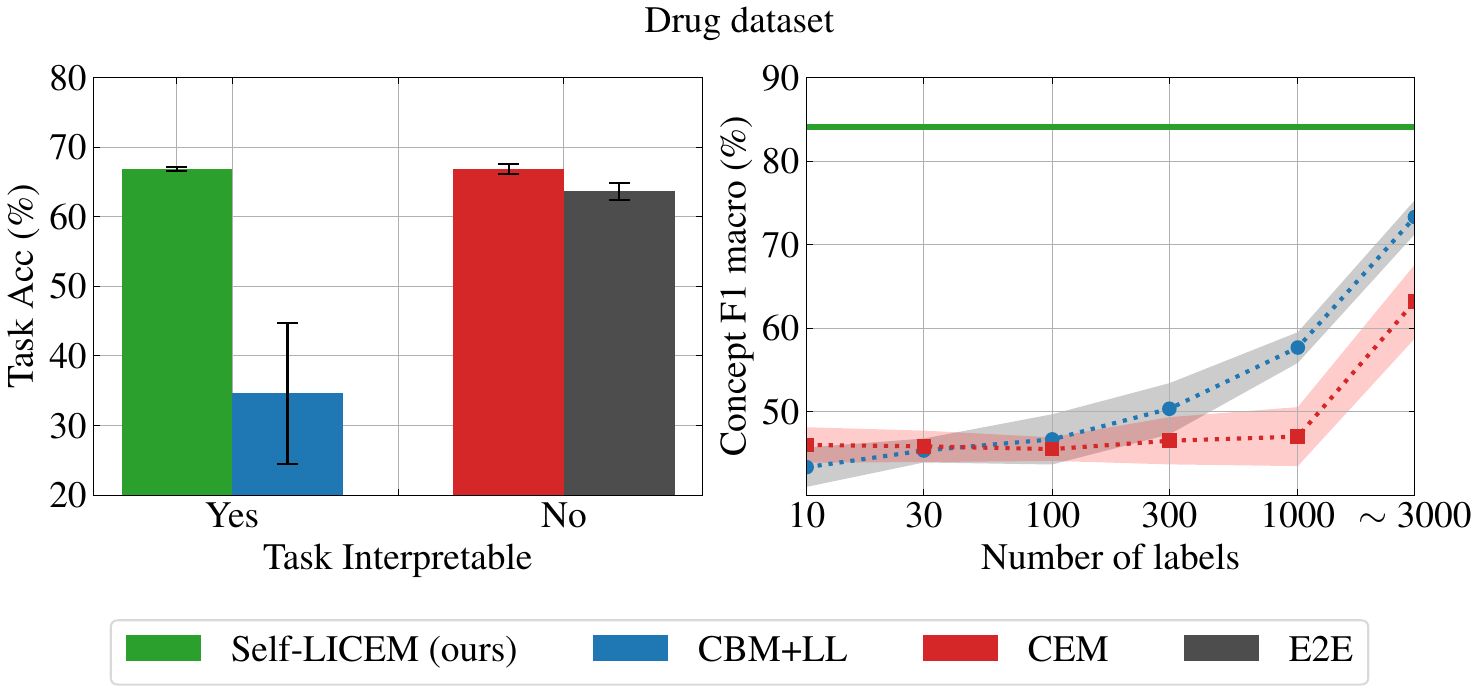}
        \caption{Left, LICEM predicting the sentiment of a drug review~\citep{grasser2018aspect}. LICEM provides accurate predictions and reveals its decision-making process.  Middle, LICEM provides the best accuracy/interpretability trade-off. Right, models' concept F1 scores, %of a Self-LICEM in comparison to standard supervised CBM and CEM, 
        when increasing the number of concept annotations. Self-LICEM achieves high scores without requiring concept labels.} 
        \label{fig:viz-abstract}
\end{figure}

This paper tackles these challenges by proposing a novel Linearly-Interpretable Concept Embedding Model (LICEM). LICEM provides the final classification through an interpretable linear equation over concepts. Specifically, both the weights and the independent variables (concept predictions) of the linear equation are predicted for each individual sample. As illustrated in Fig.~\ref{fig:viz-abstract} (left), LICEM identifies a few relevant concepts in the text: an \textit{Effective} drug with \textit{Side Effects}. While the presence of \textit{Side Effects} normally indicates a poor drug, LICEM adjusts the concept relevance according to the meaning of the specific input text. In this case, LICEM recognizes that the predicted side effect is marginally relevant because ``most other medications result in [it],'' thus assigning low concept importance to \textit{Side Effects} while giving higher importance to \textit{Effective}, leading to an overall positive review for the drug. 
In contrast, the CBM provides an incorrect prediction because it relies solely on the predicted concepts and their global importance for predicting a given task. In the reported example, CBM predicts a negative drug review solely due to the presence of \textit{Side Effects}. Furthermore, when employing a non-linear task predictor, the CBM decision-making process is non-interpretable from the concept to the task, while LICEM explicitly reveals its reasoning.

In the experiments, we show that LICEM addresses all the above-mentioned CBM issues. In particular, we show that: i) LICEM achieves higher accuracy than existing task-interpretable models (e.g. CBM+LL, a CBM with a linear layer on top) while matching or surpassing black-box methods (Fig.~\ref{fig:viz-abstract}, middle); 
ii) LICEM explanations are more intervenable and causally consistent with respect to existing solutions; %plausible and useful with respect to existing solutions by means of a user study; 
iii) LICEM can be trained without any concept annotation (Self-LICEM), as concepts can be automatically predicted by its LLM backbone, often providing higher concept accuracy than standard methods with full annotations %an existing method 
(Fig.~\ref{fig:viz-abstract}, right).

\section{Related Work}
\label{sec:rel_work}
This section reviews prior work relevant to our approach, with a focus on LLM interpretability and concept-based models.

\textbf{LLM interpretability. }
Recent studies have highlighted the unreliability of LLMs, as they often occur hallucinations~\citep{ji2023survey}, and when prompted for explanations, their responses frequently do not reflect the actual decision-making process~\citep{ye2022unreliability, madsen2024self, turpin2024language}. Although the attention mechanism in transformer models offers some interpretability, it has been criticized for its lack of clarity and consistency~\citep{jain2019attention, wiegreffe2019attention}. %This underscores the urgent need to clarify LLM decision-making processes, as it hinders their adoption in fields where transparency is crucial~\cite{gptlimits, ziems2024large}.
To improve LLM explainability, various standard XAI techniques, such as LIME~\citep{ribeiro2016should} and Shapley values~\citep{lundberg2017unified}, along with newer methods~\citep{bertMeetsSHapley, heyen2024effect, chefer2021transformer, chefer2021generic}, have been employed. However, these standard techniques have  limitations~\citep{kindermans2019reliability, ghorbani2019interpretation, adebayo2018sanity, taimeskhanov2024cam}, primarily because they explain predictions in terms of input features that often lack meaningful interpretations for non-experts~\citep{poursabzi2021manipulating}. Consequently, researchers are now exploring interpretable-by-design models also in the textual domain~\citep{rajagopal2021selfexplain, jain2022extending, tan2024interpreting, tan2024sparsity}.
%While the rise of LLMs has brought about fresh challenges in XAI, it also presents opportunities, as

\textbf{Concept-based models. }
Concept-based models~\citep{alvarez2018towards, koh2020concept, ciravegna2023logic, kim2023probabilistic} are transparent and interactive models that utilize an intermediate layer to represent concepts. % enabling human experts to intervene and extract counterfactual predictions. 
%However, the concept layer can create a representation bottleneck, particularly when few concepts are available. 
To increase the representation capability of the concept layer, \citet{zarlenga2022concept} proposed using concept embeddings. %—rich vector representations that alleviate the information bottleneck~\citep{shwartz2017opening}. 
However, the interpretability of CEM task predictor is limited, as individual embedding dimensions lack clear meaning. In this work, we demonstrate how to create an interpretable task predictor over these embeddings. A recent neurosymbolic method (DCR,~\cite{barbiero2023interpretable}), based on fuzzy logic, also attempted to tackle this issue. While we extend CEM and DCR applicability to the textual domain, we show that LICEM achieves superior predictive performance than DCR and higher interpretability than both, as confirmed by a user study. 
Additionally, supervised concept-based models~\citep{koh2020concept, zarlenga2022concept} often require extensive concept annotations, which are frequently unavailable, particularly in text. We enhance a recent generative approach~\citep{yang2023language, oikarinen2023label, ludan2023interpretable} by using the same LLM for self-generated concept predictions and sample representations.

\section{Background}
\label{sec:background}
Before introducing our proposed methodology, we outline the foundational methods that underpin our study. Specifically, we first describe CBM and CEM and then explain how Large Language Models (LLMs) can be leveraged to extract rich textual representations. %We will review relevant literature on CBMs and the challenges associated to efficiently exploit existing knowledge in LLMs.  %
%This foundation enables us to contextualize the proposed approach and highlight the limitations of existing solutions.
 
\textbf{CBMs. } 
As shown in Figure~\ref{fig:viz-abstract} (left),  CBMs~\citep{koh2020concept, tan2024sparsity} are transparent models that break the standard end-to-end learning paradigm into the training of two neural modules $f \circ g$. The concept encoder $g: X \rightarrow C$ maps raw features $x \in X \subset \mathbb{R}^d$ into $m$ higher-level abstractions $c \in C \subset [0, 1]^m$ (i.e., the concepts); the task encoder $f: C \rightarrow Y$ predicts $n$ downstream classes based on the learned concepts $\hat{y} = f(g(x)), y \in Y \subset [0, 1]^n$. 
Recalling the example in Figure~\ref{fig:viz-abstract}, a CBM decomposes the classification task into an inital prediction of drug-related attributes -- the presence of \textit{Side Effects} and its \textit{Effectiveness} --  which are subsequently used to evaluate the overall drug sentiment.
CBMs are normally trained to minimize a composite cost function, considering concept and task learning: $ \mathcal{L} = H(\hat{c}, c) + \lambda \cdot H(\hat{y}, y)$, where $H$ denotes the standard cross-entropy function and $\lambda\in[0,1]$ is a coefficient used to prioritize concept learning relative to task learning.
CBMs are considered more interpretable than DNNs because they employ a transparent intermediate representation, and they inherently deliver counterfactual predictions. %and expose the relations between concepts and classes when a single linear layer is utilized as task predictor.
However, the expressivity of CBM is limited by the bottleneck representation created by the concept layer. This issue is particularly relevant when a single layer is used as task-predictor and when the concepts employed are incomplete~\citep{yeh2020completeness}, i.e., they are not sufficient to uniquely distinguish the final classes, % In other words, CBM performance is limited in scenarios where the set of concepts is incomplete for the task at hand~\citep{yeh2020completeness}, 
a frequent condition in Natural Language Processing (NLP) contexts. Figure \ref{fig:viz-abstract} shows a case where this limitation is evident. While \textit{Side Effects} are usually linked to negative sentiment, they are irrelevant in this example. Yet, the CBM still predicts a negative outcome, as it relies solely on the concept predictions and is unable to capture the broader context.

\textbf{CEMs. }
 Concept Embedding Models (CEMs)~\citep{zarlenga2022concept, kim2023probabilistic} address the limited expressivity of CBMs by generating a concept embedding to represent each concept. 
 Initially, CEMs decompose the concept encoder into two functions $g = q \circ h$. The inner function $h: X \rightarrow H \subset \mathbb{R}^b$ provides a representation of an input sample, while $q: H \rightarrow \mathbf{C}$ maps this representation into $m$ $k$-dimensional concept embeddings $\mathbf{c} \in \mathbf{C} \subset \mathbb{R}^{m, k}$. The concept prediction $\hat{c}_j$ is then given by a neural function over the concept embeddings $\hat{c}_j = s(\mathbf{c_j})$, where $s$ is shared among the $m$ concepts, %A probabilistic formalization can be found in Appendix~\ref{app:stat_formalization}. 
while the task prediction is generated by a task  function $f: \mathbf{C} \rightarrow Y$ $f(\mathbf{c})$ working on the concatenation of all concept embeddings. Hence, as shown in Figure~\ref{fig:viz-abstract} (middle) the expressivity of CEM is much higher than CBM, as it is not constrained to represent concepts with single neurons.
sOn the other side, the interpretability of the CEM task predictor $f(\mathbf{c})$ is very limited, as the individual dimensions of concept embeddings are not interpretable: while the concept $c_\text{Side Effects} = 0.7$ is interpretable, the single dimensions associated to the corresponding embedding $\mathbf{c}_\text{Side Effects} = [0.2, 2.5, \ldots, -1.7]$ are not semantically meaningful. CEM makes predictions on top the concept embeddings, thus, even when using a single linear layer, the task predictor is not interpretable. Ideally, we want a task predictor that combines the expressiveness of CEM with the interpretability of logistic regression applied directly to concept predictions.
Also, to date, no adaption of CEM architectures to text scenarios has been proposed.

\textbf{LLM-based Textual Encoders. }
When considering transformer models, there exist several methods for implementing a text encoder $h(x)$. An immediate choice is to employ an encoder-only architecture, such as BERT~\citep{devlin2018bert}, and extracting the embedding associated to the [CLS] token. However, as recently shown in~\citet{jiang2023scaling}, one can also exploit the remarkable performance of existing decoder-only LLMs. The architecture of an LLM can be conceptualized as comprising two distinct components: the stacked decoder blocks which are responsible for generating a contextualized representation $e$, and a classification head that processes this representation to predict the next token. The stacked decoder blocks can be formalized as a function $h(\cdot):D^l\rightarrow \mathbb{R}^b$, where $D$ denotes the dictionary of tokens recognized by the LLM, and $l$ represents the length of the context window. This function maps an input $x$ to its vector representation $e = h(x)$. %which the classification head then associates with the token predicted to continue the prompt $t$. 
To facilitate the generation of sentence-representative embeddings, \citet{jiang2023scaling} proposed embedding the input within the prompt as \textit{``this sentence: [sentence] means in one word: "} and substituting \textit{``[sentence]"} with the specific text to encode. % to be embedded, which showed to be 

%In order to obtain a rich representation of the $x$ sequence, exploiting the available knowledge in the pretrained LLM, we thus use the embedding $e \sim p_h(e|t,x)$.

%%% CONSIDER MOVING HERE THE GENERATIVE APPROACH 

%To represent an entire input $x$ one can require the model to provide a description of an input sentence by means of a single token. %\todo{ANCHE QUI}. 
%In other words, one can prompt the LLM with \textit{"this sentence: `[sentence]' means in one word: "}, and then use the subsequently generated embedding to represent the provided sentence.
%and then use the embedding of the subsequently generated token to represent the provided sentence.

%LLMs are generative methods that \todo{PIETRO e FRANCESCO aggiungere qui descrizione probabilistica}.

\section{Method}
\label{sec:methods}

\begin{figure}
    \centering
    \includegraphics[width=\textwidth]{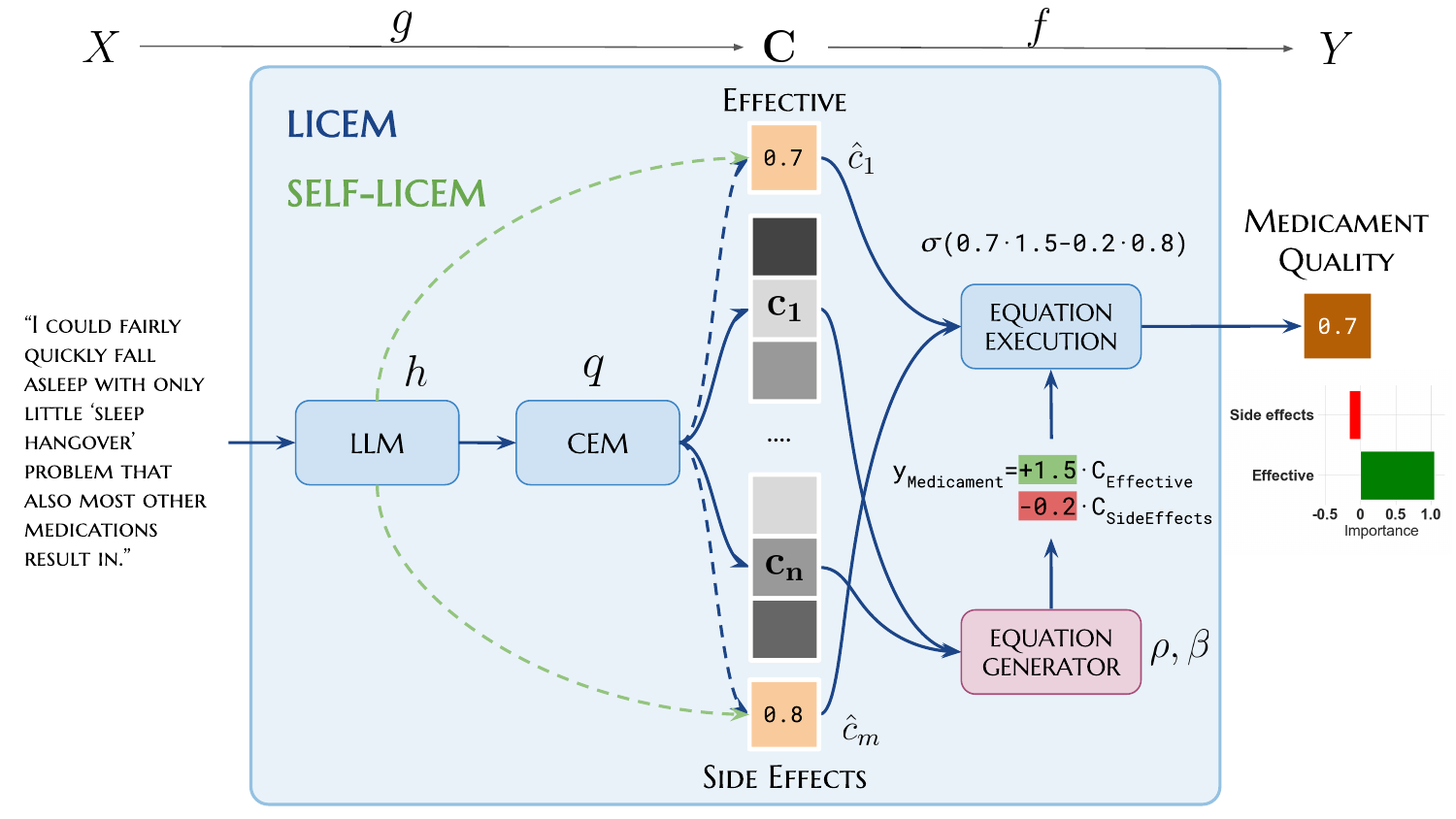}
    \caption{LICEMs visualization. Using a pretrained LLM model, we i) require it to provide an encoding of the input text following; ii) prompt the LLM to generate the concepts predictions $\hat{c}$ (e.g., Side Effects = 0.8) in Self-LICEM, while in LICEM they are provided by a concept embedding layer; iii) make the final prediction in an interpretable way by first predicting the equation weights $w_{ij}$ (e.g., $w_{\text{Side Effects}}=-1.8$)  for predicting the $i$-th class, then executing the resulting linear equation.}
    \label{fig:pipeline}
\end{figure}

In this paper, we propose an interpretable concept-based model for text classification that leverages rich text and concept representations.  Fig.~\ref{fig:pipeline} illustrates the complete pipeline. A pretrained LLM is used as the text encoder to extract contextualized representations by employing the prompting strategy introduced in~\cite{jiang2023scaling}, thus generating the embedding $e=h(x)$ for the input text without requiring fine-tuning of the encoder. The encoded text is then passed through a concept embedding layer~\cite{zarlenga2022concept}, producing concept embeddings $\mathbf{c} = q(e)$ and corresponding concept predictions $\hat{c} = s(\mathbf{c})$. The proposed model (LICEM, Section~\ref{sec:licem}) produces an interpretable task prediction by leveraging both the concept embeddings and predictions.%, while also revealing its internal decision-making process. 
Furthermore, using a pretrained LLMs as text encoder allows for the self-generation of concept predictions (Self-LICEM, Section~\ref{sec:selfsup}). %, extending the applicability of CBMs to scenarios where concept annotations are unavailable.

%In Section~\ref{sec:icem} we describe LICEM, a novel linearly-interpretable task predictor working over concept embeddings, while in Section~\ref{sec:selfsup} we propose a self-generative approach for training LICEMs in scenarios where concept annotations $c$ are not available (Self-LICEM). 
%finally, in Section~\ref{sec:control}, we propose to guide the generation process of an LLM by means of an I-CEM interpretable function, ensuring its adherence to the provided decision-making path.

\subsection{Linearly-Interpretable Concept Embedding Model (LICEM)}
\label{sec:licem}
To create an interpretable model, it is essential to utilize both an interpretable data representation and an interpretable function~\citep{ribeiro2016should, rudin2019stop}.
To avoid generalization losses, state-of-the-art concept-based models lacks one of the characteristics: they either employ non-linear functions on top of concept predictions (losing functional interpretability), or any function on top of concept embeddings whose single dimensions are non-interpretable (losing data interpretability). 

To address this issue, in this work, we propose to \textit{neurally generate a linear equation} that can be \textit{symbolically executed} over the concept predictions. In this way, the final classification is outputted as an interpretable aggregation of the most important concepts. 
More precisely, we predict the weights and the bias of a  set of linear equations each one predicting one class, and in which the independent variables are the concepts scores $\hat{c_j}$ predicted by CEM concept encoder. The prediction of both parameters (weights and biases) is provided by two neural modules $\rho$, $\beta$ working over CEM concept embeddings $\mathbf{c}$. The first neural module $\rho: \mathbf{C_j}\rightarrow \mathbb{R}^{n}$ predicts for a single concept $j$ the weights for all classes $\hat{w}_{j} = [\hat{w}_{1j}, \ldots, \hat{w}_{ij}] =  \rho_i(\mathbf{c_j})$. 
The second neural module $\beta: \mathbf{C} \rightarrow \mathbb{R}$ predicts the bias term $\hat{b} = [b_1, \ldots, b_n] = \beta(\mathbf{c})$ over the concatenation of all concept embeddings $\mathbf{c} = [\mathbf{c_1}, \ldots, \mathbf{c_m}]$, which represents the general bias for each class.
%Notice that also the concept values are also predicted, although by CEM concept encoder. %Specifically, we employ two neural modules to predict \textit{for each sample} the weights and the bias of a linear equation that is executed over the concepts (that are also predicted). 
Overall, we can describe LICEM predictions as:
\begin{equation}
    \text{LICEM}: \quad \hat{y_i} = \sigma\left(\sum_j \hat{w}_{ij} \hat{c}_j  + \hat{b}_i\right) %\qquad \hat{w}_{ij} = \rho_i(\mathbf{c_j}), \ \hat{b}_i = \beta_i(\mathbf{c}), \ \hat{c}_j = s(\mathbf{c_j}) 
    \label{eq:licem}
\end{equation}
where, as in common logistic regressions, $\hat{w}_{ij}$ is the weight for the $j$-th concept in predicting the $i$-th task, $b_i$ is the bias for the $i$-th task, $\hat{w}_{ij} < 0$ indicates a negatively important concept, $\hat{w}_{ij} > 0$ a positively important one, and $\hat{w}_{ij} \sim 0$ a non-important concept. 
Notice that the bias term is optional, but it allows for positive predictions even when no concept is positively predicted. Indeed, when $\hat{c}_j = 0$ for all $ j \in \{1,...,m\} $, the prediction would be $\hat{y}_i=0$ regardless of $\hat{w}_{ij}$. 
Also, $\sigma$ represents a sigmoid activation function for binary classification tasks and a softmax for multi-class classification tasks.

 %Although not experimented in this paper, the proposed model can be employed also for regression tasks by simply employing an identity function as $\sigma$. 
Finally, to understand the contribution of a concept to the final prediction of a class, %one can look at the predicted weight $\hat{w}_{ij}$ or the associated predicted concept $\hat{c}_{j}$. However, since the importance of the concept is determined by both terms, 
we propose considering the combined contribution $\hat{w}_{ij}\hat{c}_j$ and reporting them %plotting them in a LIME-like 
with a feature importance plot, as shown in the output of Fig.~\ref{fig:pipeline}.

\textbf{Training. }
LICEM is trained similarly to standard concept-based models, with a cross-entropy $H$ loss over the predicted concepts and tasks, with the addition of a few regularizations. To improve the readability of LICEM equations, we add an $ L_1 $ regularization promoting sparse weights (i.e., $\hat{w}_j \neq 0$), thus equations composed of few terms. The analysis regarding the sparsity achieved by LICEM is shown in Appendix~\ref{app:sparsity}. To prevent over-reliance on the bias term, we also add an $L_2$ regularization to encourage small bias values and minimize its influence on task prediction. 
\begin{equation}    
\mathcal{L}_{sup} = H(c, \hat{c}) + \lambda_y H(y, \hat{y}) + \lambda_w ||w||_1 + \lambda_b ||b||_2  
\label{eq:loss}
\end{equation}
where we indicate the loss over the concept predictions as $\mathcal{L}_c = H(\hat{c},c)$, the loss over the task predictions as $\mathcal{L}_t = H(\hat{y}, y)$, with $||w||$ and $||b||$ the regularization terms over the weights and biases and with $\lambda_y, \lambda_w$ and $\lambda_b$ the optimization weights for each term. In the rest of the paper, we will refer to this strategy as \textit{supervised}.  

\subsection{Exploiting LLMs to avoid concept annotation: a self-generative approach}
\label{sec:selfsup}
To alleviate human annotators from the burden of providing concept supervision, a few works are starting to exploit the knowledge already available in pre-trained LLMs, both in the image~\citep{yang2023language, oikarinen2023label} and in the textual domains~\citep{tan2024interpreting}.
%As defined in~\cite{poeta2023concept}, the idea behind concept-based generative approaches is to exploit the knowledge already available in LLMs. 
First, an LLM is asked to provide several attributes that describe each class. Each attribute is considered a concept for that class, possibly shared with other classes. For instance, a \textit{parrot} may be described as having \textit{bright feathers} and \textit{medium size}. Then another LLM is required to predict whether the concept is present in the input samples. The LLM, in this case, is formally represented by the distribution $p_\theta$, where $\theta$ denotes the parameters of a pre-trained LLM with classification head. When conditioned on a prompt $t$, the model generates the token “yes” if a specific concept is identified in the input text sequence $x$, and “no” otherwise. Thus, the predicted concept is sampled as $c' \sim p_\theta(c'|t,x)$. In Appendix~\ref{app:prompts} we report some examples of prompts.  

% vecchio paper citato: ludan2023interpretable

\textbf{Generative approach. } 
In \cite{tan2024interpreting}, these concept predictions $c'$ are used as labels to train a textual concept encoder. Formally, $\mathcal{L}_{gen} = \mathcal{L}_{c'} + \lambda \mathcal{L}_t = H(c', \hat{c}) + \lambda H(y, \hat{y}).$
We will refer to this strategy as \textit{generative}, as a generative model provides concept annotations.

\textbf{Self-generative approach. }
While the generative approach reduces human annotation efforts, it requires training an additional concept encoder to learn the LLM-provided labels.  In this paper, since we already employ an LLM as a text encoder, we propose using the same LLM to directly make the concept predictions. More precisely,  we prompt the LLM to provide both a representation $e$ for each sample $x$ and the concept predictions, i.e., $\hat{c} = c'\sim p_\theta(c'|t,x)$.
This results in a modification of both CEM and LICEM as the concept predictions are self-generated by the same LLM, as shown in Fig. \ref{fig:pipeline}.
We will refer to this approach as \textit{self-generative}, as the same model directly provides the concept predictions.  
This method eliminates the need for concept annotations, but also reduces the number of parameters to train and improves concept performance if compared to the generative method. Indeed, the concept accuracy of the self-generative method represents an optimum for the generative one. In the former, the concepts $c'$ provided by the LLM are directly used as concept predictions, while in the latter, they serve as training labels for an external text encoder, which aims to replicate $c'$.  
%Indeed, as we will show in the experiments, the concept predictions produced by the LLM in the self-generative approach persist as an upper bound for the {generative} approach as they are the learning objective of the generative concept encoder.
Self-LICEM is obtained by substituting the concept predictions $\hat{c}$ with $c'$ from Equation~\ref{eq:licem}:
\begin{equation}
    \text{Self-LICEM} \quad \hat{y_i} =  \sigma\left(\sum_j \hat{w}_{ij} c'_j  + \hat{b}_i\right).
\end{equation}
The concept embedding encoder $q$ and the neural modules $\rho$ and $\beta$ producing the interpretable linear equation are trained as in Equation~\ref{eq:loss}, but minimizing, this time, only the loss over the task:
\begin{equation}
\mathcal{L}_{self gen} = H(y, \hat{y}) + \lambda_w||w||_1 + \lambda_b||b||_2,
\label{eq:loss_self}
\end{equation}
This approach is not limited to LICEM; it can also be extended to CBM-based and CEM-based models. In these cases, the LLM provides the concept predictions (CBM) or the predictions and the embedding (CEM). In both cases, the optimization strategy involves minimizing only the cross-entropy on the task predictions $H(y, \hat{y})$, as shown in Eq.~\ref{eq:loss_self}. This enables converting any pre-trained LLM into a concept-based model without the need for concept annotations.

\section{Experiments}
\label{sec:exp}
In this section, we want to answer the following research questions:
\begin{itemize}
    \item \textbf{Generalization.} Does LICEM achieve superior performance in text analysis compared to other interpretable models, and is it on par with non-interpretable ones? How does the self-generative approach perform? (Section~\ref{sec:generalization})
    \item \textbf{Concept Efficiency.} How many concept supervisions are required to match Self-LICEM accuracy? Does the self-generative strategy outperform the generative one in concept accuracy?  (Section~\ref{sec:efficiency})
    \item \textbf{Interpretability.} Can we effectively interact with LICEM? Are LICEM explanations clear and driven by most important concepts? (Section~\ref{sec:interpretability}) 
\end{itemize}

\subsection{Setup}
\label{sec:setup}
We test LICEM performance over different datasets (both with and without concept-supervisions), comparing against several models and for different metrics. For all experiments, we report the average and standard deviation across three repetitions. The models were trained on a dedicated server equipped with an AMD EPYC 7543 32-Core processor and one NVIDIA A100 GPU\footnote{Our code is available at \href{https://github.com/francescoTheSantis/Linearly-Interpretable-Concept-Embedding-Model-for-Text/tree/main}{https://github.com/francescoTheSantis}.}.

\textbf{Dataset.}
We evaluated LICEM performance on three text classification datasets with available concept annotations: {CEBaB}~\citep{abraham2022cebab}, {MultiEmotions-IT}~\citep{sprugnoli2020multiemotions}, and {Drug}~\citep{grasser2018aspect}. {CEBaB} is a dataset designed to study the causal effects of real-world concepts on NLP models. It includes short restaurant reviews annotated with sentiment ratings at both the overall review level (positive, neutral, and negative reviews) and for four dining experience aspects, which we use as concept labels: `Good Food', `Good Ambiance', `Good Service', and `Good Noise'. {MultiEmotions-IT} is a dataset designed for opinion polarity and emotion analysis, containing comments in Italian related to videos and advertisements posted on social media platforms. These comments have been manually annotated according to different aspects, from which we selected two dimensions: opinion polarity, describing the overall sentiment expressed by users (used as task label), and basic emotions. We selected `Joy', `Trust', `Sadness', and `Surprise' as the concept labels. The {Drug} dataset provides patient reviews on specific drugs. The reviews are annotated with the overall satisfaction of users (which we discretized to a binary representation) and drug experience annotations, namely `Effectiveness' and `Side Effects', which we used as concept labels.
Additionally, we tested the generative and self-generative approaches on the {Depression}~\citep{yates2017depression}\footnote{For this dataset, we used the cleaned version available on \href{https://www.kaggle.com/datasets/infamouscoder/depression-reddit-cleaned}{Kaggle}.}, IMDb~\citep{maas2011learning}, TREC-50~\citep{liroth2002learning,hovyetal2001toward}, Banking-77~\citep{Casanueva2020} and CLINC-OOS~\citep{larsonetal2019evaluation}. %This dataset consists of Reddit posts from both users who claimed to have been diagnosed with depression and control users. Since in this case concept annotations are unavailable, we prompted an LLM~\citep{jiang2024mixtral} that has identified six depression-related concepts: `Self-deprecation', `Loss of Interest', `Hopelessness', `Sleep Disturbances', `Appetite Changes', and `Fatigue'.
These datasets span a range of domains and allow for comprehensive assessment of classification performance. The number of classes varies significantly, from binary sentiment analysis in IMDb (2 classes) to fine-grained intent detection in CLINC-OOS (151 classes). This diversity ensures robust evaluation across different levels of classification complexity. Additional details regarding the datasets are reported in Appendix~\ref{app:exp_details}.

\textbf{Baselines.}
We compare LICEM against several baselines, including black-box and concept-based models, both task-interpretable and non-interpretable approaches.
For all models, we use a non fine-tuned Mixtral 8x7B~\citep{jiang2024mixtral} encoder $h(x)$, following the encoding strategy proposed in \cite{jiang2023scaling}. In Appendix~\ref{app:enc_comp} we also report all results based on a fine-tuned BERT encoder~\citep{devlin2018bert} as backbone. The results show that the decoder-only LLM achieves similar performance without fine-tuning the whole LLM. Besides, it enables the self-generative approach: in Appendix~\ref{app:annotation_comparison} we report a comparison of the concept annotation performance when using different LLMs.
For black-box models (\textsc{E2E}), we evaluate an end-to-end model directly classifying the task with a Mixtral encoder $h(x)$ and few layers as classification head (MLP), and the same Mixtral used in Zero-shot and Few-shot prompting. 
CBM+LL and CBM+MLP are the two CBMs originally proposed in~\citep{koh2020concept} and recently adapted to text in~\citep{tan2024interpreting}. They employ a concept bottleneck layer followed, the first one, by an interpretable linear layer, while the second by a non-interpretable multi-layer perceptron. 
CBM+DT and CBM+XG are respectively two CBM variants proposed in~\citep{barbiero2023interpretable}, using a decision tree and a XGBoost classifier~\citep{chen2016xgboost} on top of the concept bottleneck layer, respectively. {CBM+DT} is task-interpretable, as one can extract a decision rule based on concepts, whereas the second variant {CBM+XG} is non task- interpretable.
As described in Section~\ref{sec:background}, {CEM}~\citep{zarlenga2022concept} employs embeddings to represent concepts and enhance CBM generalization performance, but at the cost of losing task interpretability.
Finally, {DCR}~\citep{barbiero2023interpretable} is a neuro-symbolic approach designed to improve the interpretability of CEM. It generates propositional rules executed by a fuzzy system on top of concept predictions. We adapt CEM and DCR to work in the text analysis scenario, and we compare their performance against the proposed model.
For the training details regarding each model, please refer to Appendix~\ref{app:exp_details}.

\textbf{Metrics. }
We evaluate LICEM using various metrics. To assess \textbf{generalization} performance, we compute the task accuracy and the macro-averaged concept F1 score (as concept classes are highly imbalanced); for self-generative models, the macro-averaged F1 score evaluates the concept predictions directly provided by the LLM (Section~\ref{sec:selfsup}). 
To measure \textbf{efficiency}, we examine the concept F1 score of all models when increasing the number of concept annotations.
For \textbf{interpretability}, we evaluate the effectiveness of concept interventions in LICEM to enhance classification accuracy \citep{koh2020concept} and we compute the area under the accuracy gain curve for each model-dataset combination, calculated using the composite trapezoidal rule; secondly, we measure the Causal-Concept Effect (CaCE)~\citep{goyal2019explaining}, which assesses the causal relevance of concepts for task predictions; thirdly, we qualitatively report some of the equations generated by LICEM to demonstrate their clarity; lastly, we also conducted a user study to evaluate how easily LICEM's explanations can be understood from a human perspective, providing insight into their interpretability in real-world settings.

\input{new_task_acc}

\subsection{LICEM generalization}
\label{sec:generalization}
\textbf{LICEM matches black-box task performance and outperforms all task-interpretable models (Table~\ref{tab:task_acc}). }

LICEM consistently outperforms competing models across a diverse set of datasets, achieving either the highest or statistically equivalent task accuracy. In the supervised setting, LICEM is the top-performing task-interpretable model, showing a $7–19\%$ improvement over CBM variants and a $1–2\%$ gain over DCR. This performance gap widens in the generative setting, where evaluation involves more challenging tasks (e.g., classification over 151 classes in CLINC-OOS). Here, CBM-based models perform significantly worse, and while DCR outperforms CBMs, it still falls short of LICEM, with a notable $55\%$ accuracy gap on CLINC-OOS. In the self-supervised setting, overall model performance improves, including that of CBM variants and DCR. LICEM continues to lead, outperforming DCR on 6 out of 8 datasets and matching its performance on IMDb. We attribute LICEM’s advantage over DCR to its simpler and more direct classification mechanism: LICEM predicts the parameters of a linear function, whereas DCR constructs and optimizes complex fuzzy logic rules. The linear approach requires only a weighted sum, making both training and inference more efficient.

\begin{figure}
    \centering
    \includegraphics[width=0.8\textwidth]{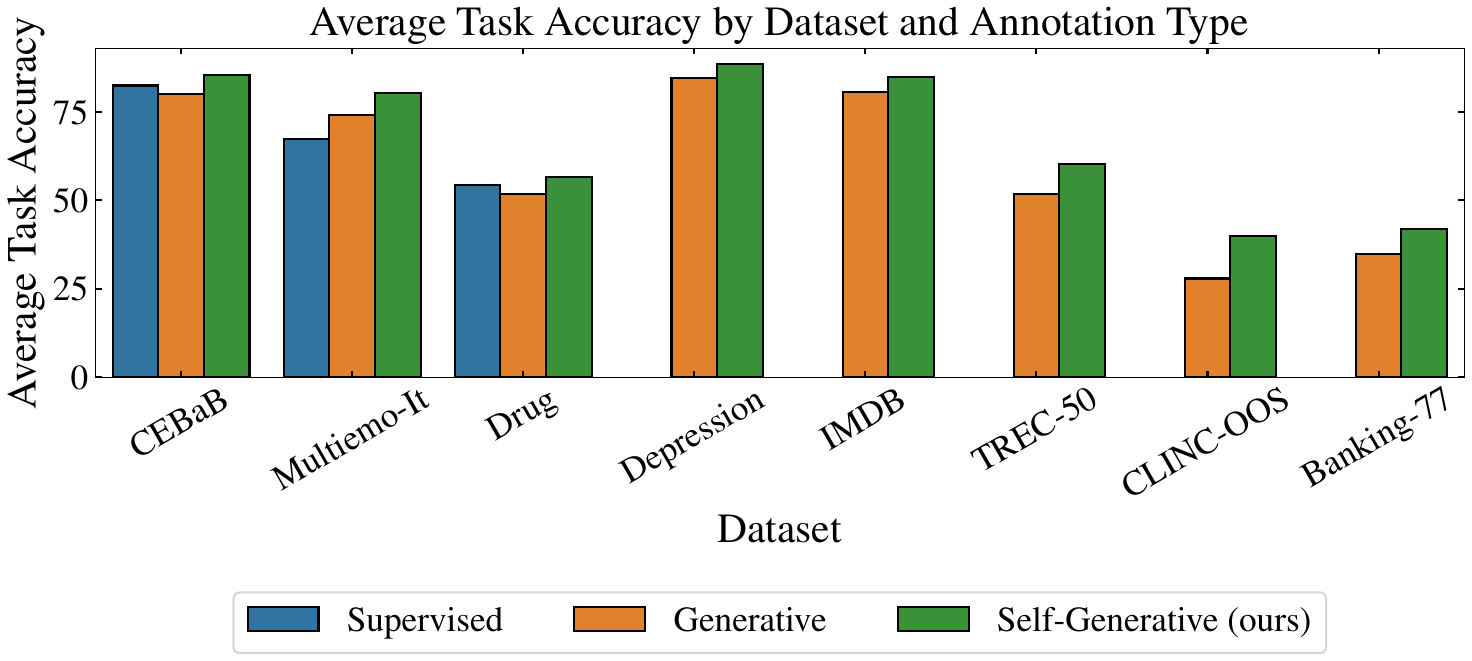}
    \caption{Average task accuracy of concept-based models grouped by annotation type over the four datasets. The self-generative approach extends concept-based models applicability to scenarios without annotations (e.g., Depression) while improving their generalization.}
    \label{fig:task_acc_comp}
\end{figure}

\begin{figure}[t]
    \centering
    \includegraphics[width=1\textwidth]{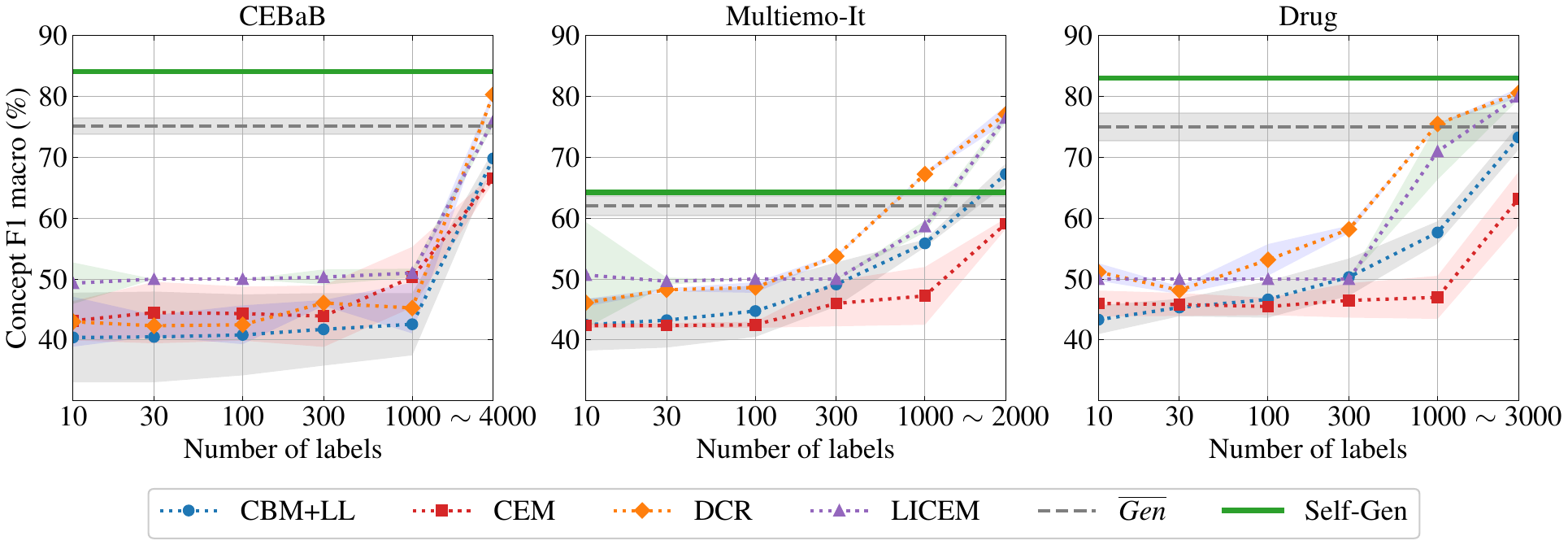}
    \caption{Concepts prediction performance vs number of concept labels used during training.  To increase plot readability, we only included the CBM+LL and the average F1 score for the generative approaches ($\overline{\text{Gen}}$). Self-generative and Gen. approaches are reported with a straight line, as they do not require concept annotation. }
    \label{fig:efficiency}
\end{figure}

\vspace{1 em}
\textbf{{Self-generative} approach increases concept-based models applicability and improves task performance (Fig.~\ref{fig:task_acc_comp}). }
To better analyze whether a certain annotation type improves the task accuracy of a model, in Fig.~\ref{fig:task_acc_comp} we report the concept-based model average task accuracies by annotation type. 
We can notice that the task accuracy of the self-generative approach is generally higher than both generative and supervised. This result is interesting because self-supervised models have fewer parameters to train than the corresponding generative ones. This behaviour is likely due to the higher concept accuracy of the self-generative approach (see Fig.~\ref{fig:efficiency}) which also affects the resulting predictions. 

%Notably, self-supervised ICEMs task accuracy matches the one of supervised models. Thus, the self-supervised concept annotation procedure does not to affect the performance of the model while it reduces the human annotation effort. 
% In comparing the model performance of self-generative, generative and supervised approaches, we can notice that the confidence intervals are overlapping. In a few cases, such as CBM+LL, we can notice a stable improvement over all the datasets when using the {self-generative} approach up to $+30\%$ on the Multiemo-It dataset. This is likely due to the trade-off posed when training a concept-bottleneck layer, which has to favor either the task or the concept performance: when directly working over good concept predictions, CBM performance improves. 
%In Appendix \ref{app:conc_acc}, we also report the task accuracy of models trained along the standard generative approach, showing similar results.
%With respect to the generative ones, the self-generative CBM variants exhibit improved performance, whereas models relying on concept embeddings show comparable results between the two approaches.
We reported the performance of all concept-based baselines (not only Self-LICEM) when trained along the {self generative} approach. This was to showcase that any concept-based model can operate on a pretrained LLM without needing concept annotations.

\subsection{LICEMs concept efficiency}
\label{sec:efficiency}

\textbf{Self-generative approach strongly reduces the human annotation effort (Fig.~\ref{fig:efficiency}). }
% \textbf{Self-supervised and generative ICEMs strongly reduce human annotation effort. }
%Self-supervised and generative models have a concept accuracy that is close to those of supervised models when using all annotations and much higher otherwise. %Even when using 1000 annotated concept labels, supervised models generally achieve a lower concept f1 score than  self-supervised and generative approaches, even though the latter do not require any human annotations. 
To assess the efficiency in terms of concept labels required to properly train the different models, in Fig.~\ref{fig:efficiency} we report the concept prediction performance when increasing the number of concept labels used for training. {Self-generative} and {generative} approaches are reported with a straight line since they do not require any concept supervision\footnote{{Generative} approaches results are reported with variance because the concepts are still learnt and thus the performance vary across models. For the {self-generative} approach, instead, the result does not vary because the concepts are predicted equally by the LLM for all models since we set the LLM's temperature to zero, % for the LLM predicting the concepts, 
which results in a deterministic annotation.}. 
Generative and self-generative models achieve a concept macro-averaged F1 score that is higher or close to that of supervised models when using all available annotations, and significantly higher otherwise.
When considering the CEBaB and Drug datasets, supervised models do not surpass self-Gen even when using all concept annotations, with the latter achieving the highest concept accuracy.
Likely, the amount of concept annotations required to match the accuracy of the self-generative approach exceeds what is available in these datasets. %(ii) the embeddings provided by an LLM may contain less information than what the same LLM can generate through self-generated predictions. 

\vspace{1 em}
\textbf{The self-generative concept accuracy exceeds that of the generative approach (Fig.~\ref{fig:efficiency}).}
The concepts prediction performance of the {generative} approach tends to be lower than that of the {self-generative} approach, with a reduction ranging from 2\% to 7\% in F1 macro score. This is because the concepts predicted by {generative} models are approximations of the self-generated concepts $c'$ used in the {self-generative} approach. These self-generated concepts serve as the labels for training the concept encoders in the {generative} learning process. For a different visualization, we report the concept accuracy when provided with all samples for all models across the three dataset with concept annotation also in Table~\ref{tab:conc_acc}, Appendix~\ref{app:conc_acc}. %{Detailed concepts prediction performance is presented}} in Appendix~\ref{app:conc_acc}, Table~\ref{tab:conc_acc} for all models across all datasets, when provided with full concept annotations.

%Self-supervised ICEMs, in particular, are capable to get a concept accuracy that supervised approaches can get close only when using all the available annotations. 

% \textbf{Self-supervised ICEMs concept accuracy is close or better than fully-supervised models. }

\begin{figure}[t]
    \centering\includegraphics[width=.75\textwidth]{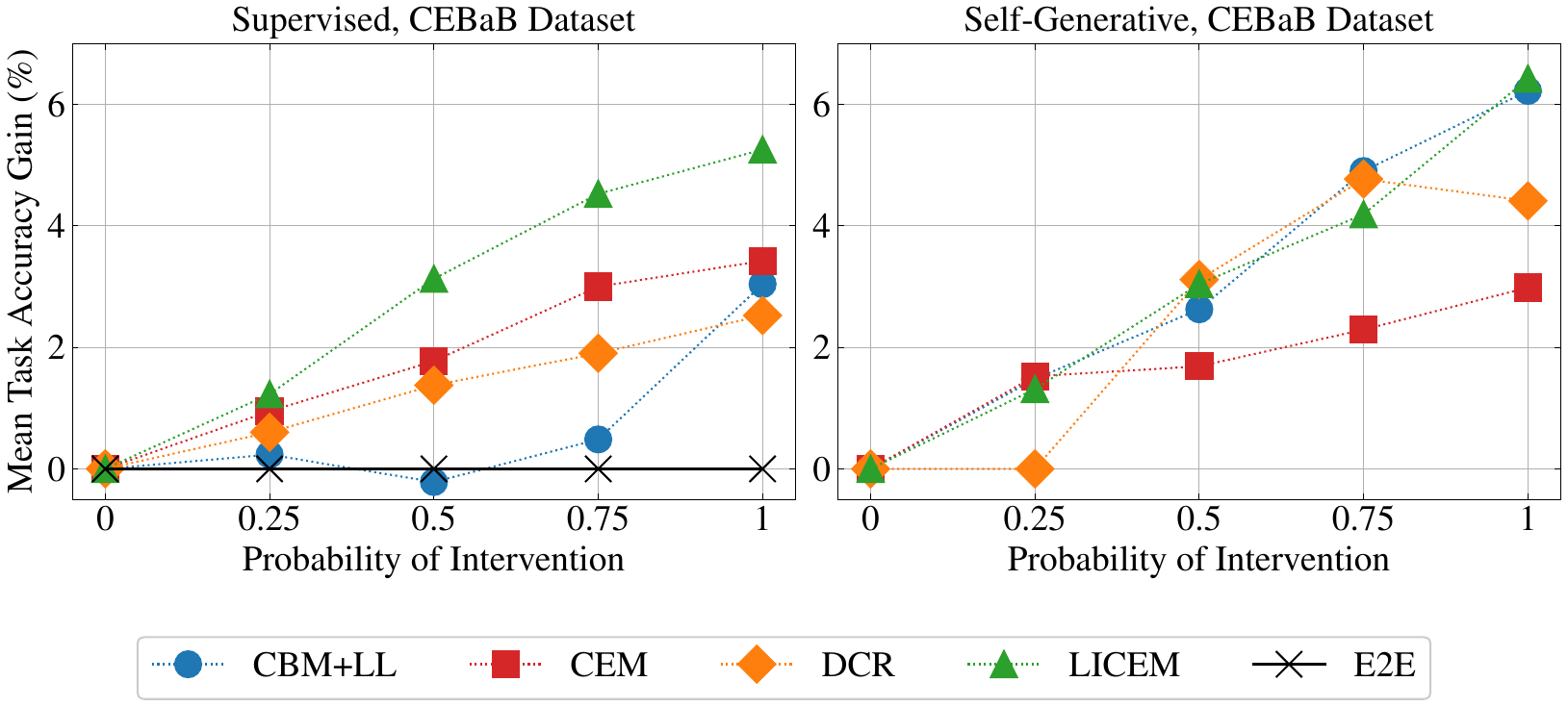}
    \caption{Concept interventions on the CEBaB dataset for (left) supervised approaches and (right) self-supervised ones.}
    \label{fig:concept_int}
\end{figure}

\subsection{LICEM interpretability}
\label{sec:interpretability}

\begin{table}[t]
{ \setlength{\tabcolsep}{16pt}
\centering
\caption{Average area under the accuracy gain curve for each model-dataset combination, calculated using the composite trapezoidal rule. The best-performing model, trained using either the supervised or self-generative approach, is highlighted in \textbf{bold}. Despite typically starting with higher accuracy, LICEM consistently improves its performance through interactions.}
\label{tab:auc_inter}
\begin{tabular}{cc|*{6}{c}}
\toprule
& Models & CEBaB & Multiemo & Drug \\
\midrule
& E2E & 0.000\tiny{ $\pm$ 0.000} & 0.000\tiny{ $\pm$ 0.000} & 0.000\tiny{ $\pm$ 0.000} \\
\midrule
\multirow{4}{*}{\rotatebox{90}{Superv.}}
&  CEM               &  1.858\tiny{ $\pm$ 0.302}            &    \textbf{10.111\tiny{ $\pm$ 0.403}}    &  2.474\tiny{ $\pm$ 0.356}           \\
&  CBM+LL            &  0.506\tiny{ $\pm$ 0.522}            &     0.973\tiny{ $\pm$ 0.368}             &  1.081\tiny{ $\pm$ 0.765}           \\
&  DCR               &  1.286\tiny{ $\pm$ 0.465}            &     4.546\tiny{ $\pm$ 0.233}             &  2.213\tiny{ $\pm$ 0.262}           \\
&  LICEM             &  \textbf{2.878\tiny{ $\pm$ 0.363}}   &     5.416\tiny{ $\pm$ 0.408}             &  \textbf{3.135\tiny{ $\pm$ 0.251}}  \\
\midrule
\multirow{4}{*}{\rotatebox{90}{Self Gen.}}
&  CEM               &      1.749\tiny{ $\pm$ 0.232}        &  2.132\tiny{ $\pm$ 0.111}                &  1.725\tiny{ $\pm$ 0.182}           \\
&  CBM+LL            &    2.878\tiny{ $\pm$ 0.082} &  4.826\tiny{ $\pm$ 0.083}                &  2.479\tiny{ $\pm$ 0.281}           \\
&  DCR               &     2.522\tiny{ $\pm$ 0.019}         &  3.299\tiny{ $\pm$ 0.036}                &  \textbf{4.090\tiny{ $\pm$ 0.048}}  \\
&  LICEM             &   \textbf{3.029\tiny{ $\pm$ 0.47}}   &  \textbf{5.416\tiny{$\pm$ 0.115}}        &  3.136\tiny{ $\pm$ 0.083}           \\
\bottomrule
\end{tabular}
}
\end{table}

\textbf{LICEM is responsive to concept interventions (Fig.~\ref{fig:concept_int}, Table~\ref{tab:auc_inter}}).
A fundamental interpretability property of CBMs is their \textit{intervenability}, i.e., the possibility to modify the concept predictions in order to correct the model or assess potential counterfactual predictions. As commonly done in CBM literature \cite{koh2020concept, zarlenga2022concept, kim2023probabilistic} we simulate this scenario by randomly replacing the concept predictions with the concept labels with increasing probability. 
%To assess the possibility to interact with LICEM, we evaluated the effect of concept interventions, i.e., modifications at test time of the predicted concepts with a concept provided by a human expert.
Fig.~\ref{fig:concept_int} shows the test task accuracy gain with increasing intervention probability on the CEBaB dataset, demonstrating LICEM's responsiveness and significant performance improvement. A similar behaviour can also be observed for CBMs, even though they were starting from a lower task accuracy and a higher increase could have also been expected. For comparison, we also report the E2E model with a flat line, since it does not offer this possibility. Concept intervention figures for all datasets are reported in Appendix~\ref{app:conc_int} showing similar results. %This result indicates that it is possible to interact with ICEMs, checking alternatives outputs when different concept predictions are suggested to the model.  %As noted in~\citep{zarlenga2022concept}, CEMs may not respond well to concept interventions, especially without conducting them during training. Thus, we trained all CEM-based models with a 0.5 intervention probability during the forward pass.
In Table~\ref{tab:auc_inter}, we summarize the performance of every model across all datasets, reporting the area under the accuracy gain curve for each model-dataset combination, calculated using the composite trapezoidal rule. The results demonstrate that, although LICEM typically starting with higher accuracy, it consistently improves its task accuracy through interactions. In four out of six scenarios, LICEM stands out as the most responsive model. When considering only task-interpretable models, LICEM is the most responsive in five cases, being surpassed by DCR only on the Drug dataset using the self-generative approach.
%In four times out of six scenarios LICEM emerges as the most responsive model. 

\begin{table}[t]
{
\setlength{\tabcolsep}{4.5pt}
\centering
\caption{Causal Concept Effect (CaCE) for different methods. A high (absolute) value implies a strong responsiveness of a model to modifications to a certain concept.}
\label{tab:cace_sup}
\begin{tabular}{ll|*{5}{c}}
\toprule
     &  Concept  &               CBM+LL    & CEM                         & DCR    &              LICEM                   &         SELF-LICEM      \\
\midrule
\multirow{4}{*}{\rotatebox{90}{CeBAB}}
& Good Food     & \g{-0.02}\tiny{ $\pm$ 0.01} & \g{0.29}\tiny{ $\pm$ 0.03}  & \g{0.33}\tiny{ $\pm0.04$}   & \g{0.62}\tiny{ $\pm$ 0.02}  & \g{0.63}\tiny{ $\pm$ 0.01} \\ 
& Good Amb. & \g{0.01}\tiny{ $\pm$ 0.05} & \g{0.08}\tiny{ $\pm$ 0.01}  & \g{0.02}\tiny{ $\pm0.01$}   & \g{0.18}\tiny{ $\pm$ 0.03}  & \g{0.20}\tiny{ $\pm$ 0.04} \\  
& Good Service  & \g{0.01}\tiny{ $\pm$ 0.04} & \g{0.13}\tiny{ $\pm$ 0.01}  & \g{0.20}\tiny{ $\pm0.08$}   & \g{0.37}\tiny{ $\pm$ 0.01}  & \g{0.35}\tiny{ $\pm$ 0.02} \\ 
& Good Noise    & \g{-0.01}\tiny{ $\pm$ 0.10} & \g{-0.05}\tiny{ $\pm$ 0.01}  & \g{-0.02}\tiny{ $\pm0.01$}   & \g{0.15}\tiny{ $\pm$ 0.02}  & \g{0.15}\tiny{ $\pm$ 0.03} \\
\midrule
\multirow{4}{*}{\rotatebox{90}{Multiemo}}
& Joy           & \g{0.04}\tiny{ $\pm$ 0.06}  & \g{0.18}\tiny{ $\pm$ 0.01}   & \g{0.16}\tiny{ $\pm$ 0.07} & \g{0.28}\tiny{ $\pm$ 0.01}  & \g{0.27}\tiny{ $\pm$ 0.01} \\ 
& Trust         & \g{0.02}\tiny{ $\pm$ 0.10}  & \g{0.60}\tiny{ $\pm$ 0.04}   & \g{0.47}\tiny{ $\pm$ 0.15} & \g{0.62}\tiny{ $\pm$ 0.03}  & \g{0.63}\tiny{ $\pm$ 0.01} \\  
& Sadness       & \g{-0.04}\tiny{ $\pm$ 0.05}  & \g{-0.06}\tiny{ $\pm$ 0.01}   & \g{-0.04}\tiny{ $\pm$ 0.02} & \g{-0.04}\tiny{ $\pm$ 0.01} & \g{-0.10}\tiny{ $\pm$ 0.02} \\ 
& Surprise      & \g{-0.01}\tiny{ $\pm$ 0.06}  & \g{0.03}\tiny{ $\pm$ 0.01}   & \g{0.06}\tiny{ $\pm$ 0.05} & \g{-0.02}\tiny{ $\pm$ 0.01} & \g{0.01}\tiny{ $\pm$ 0.01} \\
\midrule
\multirow{2}{*}{\rotatebox{90}{Drug}}
& Effectiveness & \g{0.02}\tiny{ $\pm$0.10}   & \g{0.43}\tiny{ $\pm$0.02}    & \g{0.28}\tiny{ $\pm$0.02} & \g{0.45}\tiny{ $\pm$0.04}   & \g{0.46}\tiny{ $\pm$ 0.02} \\  
& Side Effects  & \g{-0.07}\tiny{ $\pm$0.14}   & \g{-0.52}\tiny{ $\pm$0.01}   & \g{-0.25}\tiny{ $\pm$0.02} & \g{-0.55}\tiny{ $\pm$0.06}   & \g{-0.55}\tiny{ $\pm$ 0.03} \\  
\bottomrule
\end{tabular}
}
\end{table}
     
\textbf{LICEM predictions are caused by most important concepts  (Table~\ref{tab:cace_sup}). }
%% As anticipated in Section~\ref{sec:interpretability}, Concept-based models predictions must be causally influenced by the predicted concepts. We assess concept-based models' responsiveness to \textit{do-interventions} using the Causal Concept Effect (CaCE)~\citep{goyal2019explaining}, which measures the impact of input modifications on model predictions. Higher absolute CaCE values indicate stronger conditioning on relevant concepts. Tables~\ref{tab:cace_sup} shows that both supervised and self-supervised LICEM have higher CaCE values compared to CBM, CEM and DCR, suggesting stronger reliance on predicted concepts. %, while low CaCE values for CBM. 
% This result is positive since all concepts considered in this work are relevant for the task at hand. We leave for future work the exploration of tasks where there are confounding concepts and checking whether LICEM is capable to not consider them. 
In order to globally explain the task prediction of the compared models, we assess the effect to \textit{do-interventions} over concepts~\citep{pearl2016causal}, by computing the Causal Concept Effect (CaCE)~\citep{goyal2019explaining}. CaCE measures the impact of modifying input samples on model predictions. For concept-based models, interventions can be made at the concept level~\citep{dominici2024causal}. In the evaluated dataset, several concepts are globally relevant for the classification task (positively or negatively), thus we expect models to exhibit high absolute CaCE values for those values. %Therefore, models should exhibit high absolute CaCE values for these concepts, indicating that the models rely on the concept values rather than on other spurious information (e.g., concept leakage~\citep{marconato2022glancenets}). 
In Table~\ref{tab:cace_sup}, we present the results across all annotated datasets. %: both LICEMs demonstrate high scores. 
For the CEBaB dataset, the concepts of `Food' and `Service' emerge as the most crucial, while `Joy' and `Trust' hold more importance in the Multiemo dataset. %In most cases, both LICEMs reports the highest scores. 
On average, both LICEM's CaCE values are higher or on par with those of CEM, and consistently surpass those of DCR and CBM. For models using concept-embedding models (CEM, DCR), these results indicate that LICEM correctly relies more on the concept scores rather than on the concept embeddings for prediction. Thus, LICEM should be less affected by concept leakage issues~\citep{marconato2022glancenets}, and thus results more interpretable. In contrast, the low CaCE values for CBM+LL indicate a poor understanding of the task and confirms the underfitting issues outlined in Section~\ref{sec:generalization}, % a poor understanding of the task, 
likely due to the concept bottleneck representation.

\begin{figure}[h]
    \centering
    \includegraphics[width=0.495\textwidth]{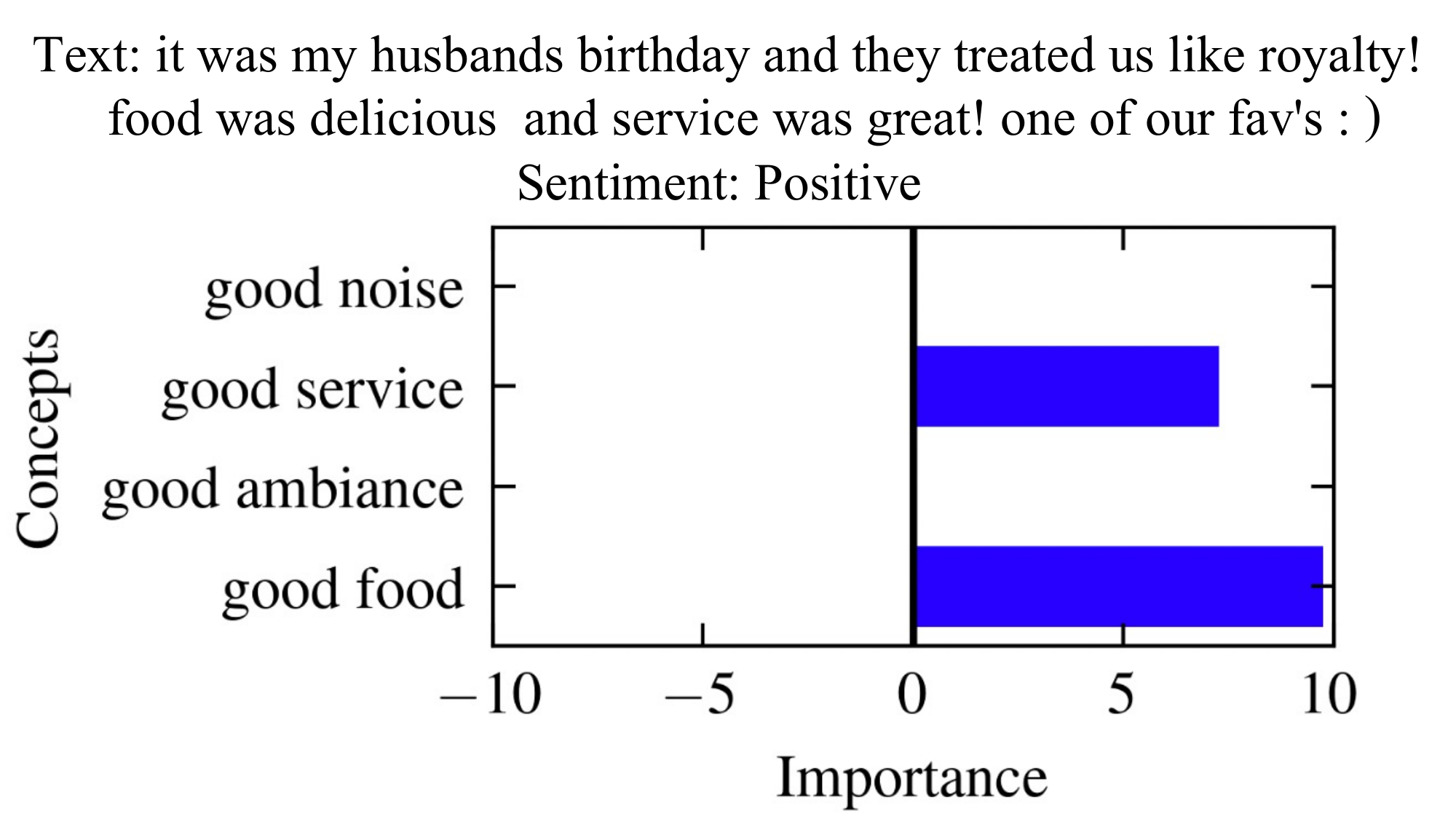}
    \includegraphics[width=0.495\textwidth]{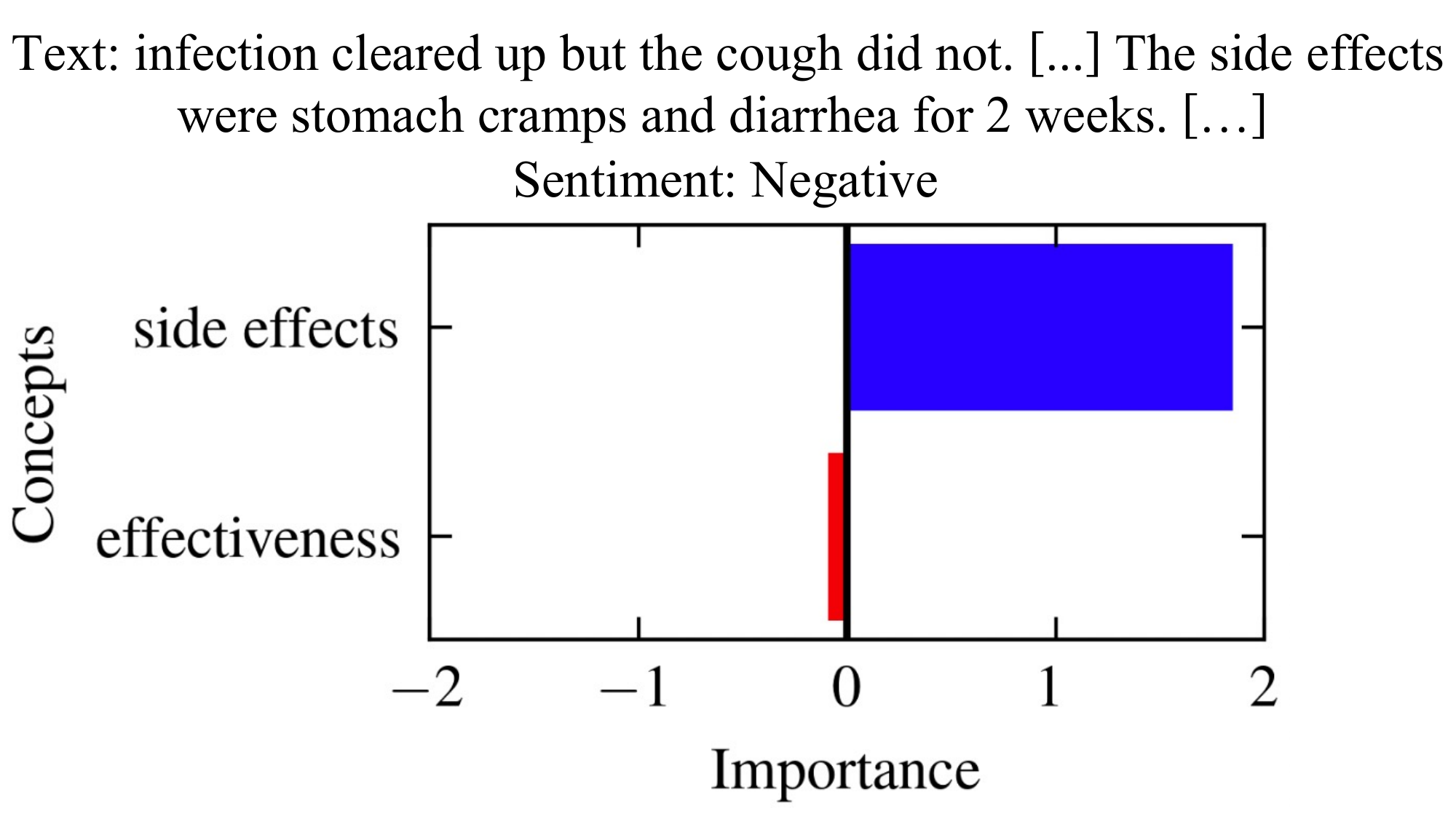}
    \caption{Explanations generated by LICEM for the predicted classes on two datasets: CEBaB (left) and Drug (right). The x-axis shows the importance scores, computed as $\hat{w}_{pj}\hat{c}_j$, where $p$ indicates the index of the class predicted by LICEM, and $j$ represents the concept index. The y-axis displays the concept names. Above the LICEM explanations, the original text and the corresponding LICEM predictions are shown. For these samples, LICEM's predictions aligned with the ground-truth values.}
    \label{fig:explanations}
\end{figure}

\textbf{LICEM predictions can be directly interpreted
%explanations are easily interpretable  
(Fig.~\ref{fig:explanations}). } 
To disclose the decision-making process underlying the prediction of a LICEM, we plot the logits $\hat{w}_{pj}\hat{c}_{pj}$ calculated by the model for the predicted class $p$ and the various concepts $j$. This visualization enables users to assess the importance of concept $j$ in the classification of the predicted class. Fig.~\ref{fig:explanations} illustrates the explanations generated by LICEM for two samples: one from the dataset CEBaB (left) and one from Drug (right). In the CEBaB example, \textit{good noise} and \textit{good ambiance} did not influence the \textit{Positive} sentiment prediction due to their absence ($\hat{c}_{pj}=0$). In contrast, \textit{good service} and \textit{good food} positively impacted the prediction, with \textit{good food} being most significant, reflecting its importance in restaurant evaluations. The Drug explanation depicts a review where the medication provided minor relief from the condition (\textit{infection cleared up but the cough did not}) but caused severe \textit{side effects} (\textit{stomach cramps and diarrhea for 2 weeks}). LICEM predicts the sentiment of the review as \textit{negative} and identifies the presence of both \textit{side effects} and \textit{effectiveness}. Since the explanation shown corresponds to the negative sentiment class, it highlights that the weight associated with \textit{side effects} is positive and substantially larger in magnitude than the weight associated with \textit{effectiveness}. Additional explanations are reported in Appendix~\ref{app:explanations}.

\begin{figure}[h!]
    \centering
    \includegraphics[width=0.8\textwidth]{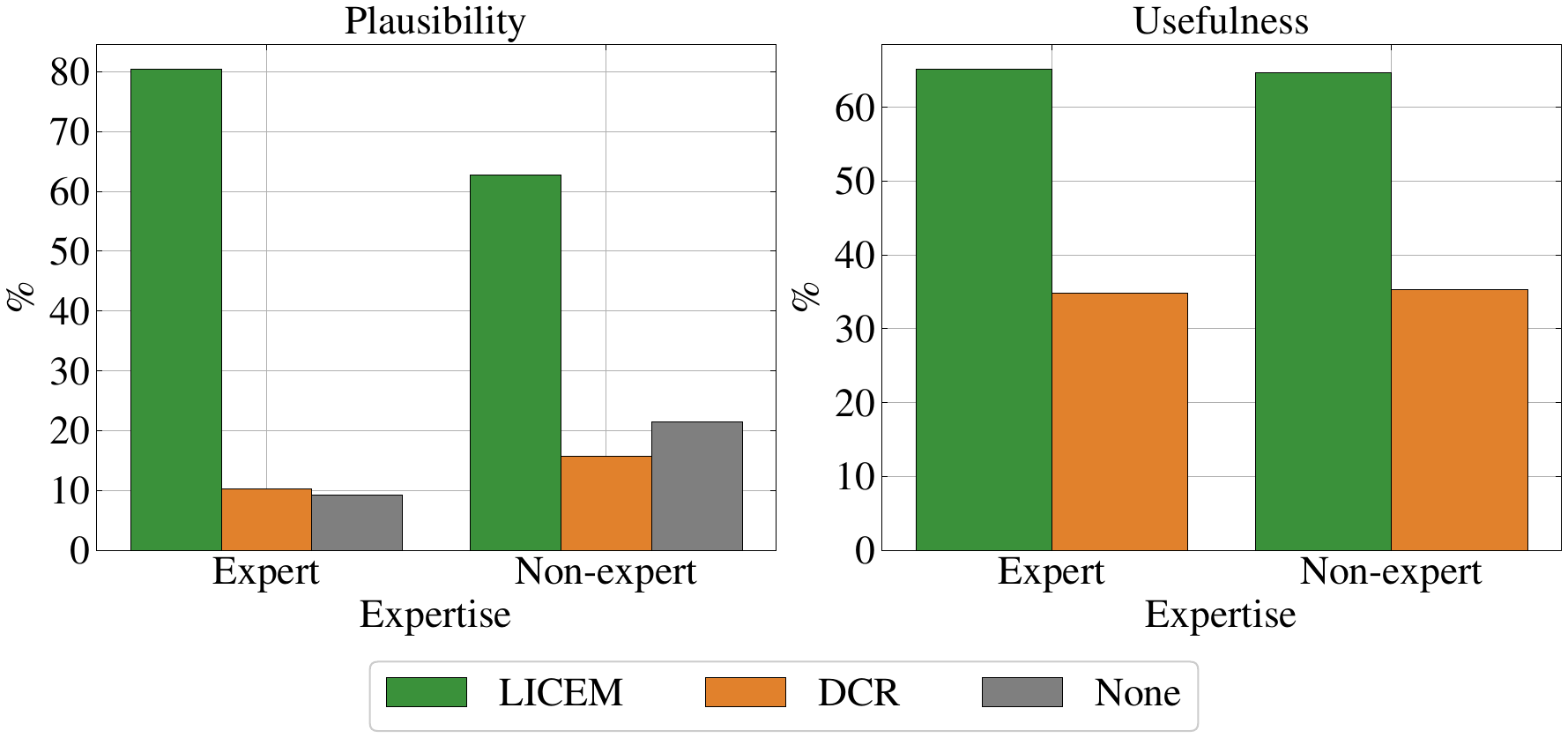}
    \caption{Averaged survey results for the two user groups. Left: explanation plausibility; Right: prediction accuracy based on explanations.}
    \label{fig:survey}
\end{figure}

\textbf{LICEM Explanations Align Better with Human Intuition (Fig.~\ref{fig:survey})}.

To evaluate the interpretability of LICEM explanations, we conducted a user study comprising $7$ questions and involving $46$ participants, consisting of both machine learning experts and non-experts (see Figure~\ref{fig:survey_info}). We compared LICEM's explanations against those generated by DCR, the strongest task-interpretable baseline. In the first task, participants are asked to choose the most plausible explanation~\citep{rajagopal2021selfexplain} from three options: the LICEM explanation, the DCR explanation, or neither. 
The explanations are extracted by randomly sampling the CEBaB dataset. In the second task, we evaluated the usefulness of explanations by measuring the participant's ability to infer the prediction of the model from the explanation provided~\citep{fel2023craft}. Examples of the two types of questions are shown in Fig.~\ref{fig:cebab_preference} and~\ref{fig:cebab_prediction_licem}. Additional explanations are shown in Appendix~\ref{app:explanations}.

The left graph of Figure~\ref{fig:survey} presents the results related to explanation plausibility. 
It is evident that the LICEM explanation is consistently considered more plausible over the rule-based DCR explanation by both expert and non-expert users. Contrary to our expectations, LICEM was especially favored by expert users, with nearly $80\%$ of them appreciating its explanations. The graph on the right of Figure~\ref{fig:survey} illustrates the accuracy achieved by the users when tasked with selecting a class label based on a given explanation. 
Both groups of users demonstrated good accuracy when making classifications using the LICEM explanations.

\section{Conclusion}
\label{sec:conclusion}
In this paper, we propose LICEM, a novel linearly interpretable concept-based model for text analysis. The experimental results show this model matches black-box models performance, is interpretable and can be trained without concept supervision (Self-LICEM). 
Besides a technological impact, we believe this work can also positively impact the society by enhancing LLM transparency and interpretability, thus facilitating their employment in several fields such as Healthcare, Finance, Legal Systems, and Policy Compliance.
As an example, in the latter case LICEM could be used to classify social media contents. LICEM would not only identify harmful posts but would also explain its decisions using concepts (e.g., hate speech, harassment) and how they contribute to the final prediction, helping moderators act quickly and transparently.
%However, LICEM could be also misused to guide LLMs to produce factually incorrect statements, thanks to their interpretability.
%Although we claim interpretability through decision rules or linear functions~\citep{rudin2019stop}, we did not perform a user study yet to confirm ICEMs ease of use. 

\textbf{Limitations. }
While LICEM improves interpretability in text classification, it still face a few challenges. First, in scenarios with a high number of concepts or fine-grained classes, the resulting linear explanations may induce cognitive overload, reducing their practical usefulness. Second, while its linear form enhances transparency, it may fail to represent complex interactions between concepts that influence task outcomes. Third, LICEM's self-generative approach hinges on the quality of concept extraction by the underlying LLM, which can vary across domains and be sensitive to prompt phrasing. %Additionally, in cases where the self-predicted concepts are noisy or semantically weak, both interpretability and accuracy may suffer. 
Finally, even with regularization, model predictions can be biased in inputs lacking clear concept signals, especially under ambiguity or domain drift.

\textbf{Future work. }
In this analysis, we focused on binary or ternary sentiment analysis for the ease of identifying concepts, and to texts composed of a few sentences.
In future work, we will extend our analysis to other NLP tasks and to longer texts, to ensure the scalability of this approach. Furthermore, we plan to extend the capability of this model to work in language modelling tasks, similarly to~\cite{ismail2023concept} employing CBMs to solve generative tasks in computer vision.  
%Furthermore, other interpretable functions could be generated and used to provide an interpretable prediction, besides linear equations. %As an example, we could also generate a text describing how each concept has been predicted and its role in the final prediction, together with the indication of the task prediction.
We leave these investigations for future research.

\bmhead{Supplementary information}
This article includes a technical appendix with supplementary details on the methodology and experimental campaign, which are helpful but not essential to the main content.

\bibliography{main}

\clearpage

\backmatter

\begin{appendices}

\section{}

\subsection{Prompts for annotation}
\label{app:prompts}
Here we report the prompts used to instruct \textit{Mistral 7B} and \textit{Mixtral 8x7B} to perform the annotations on the 4 different datasets used in this work. We adopted the in-context instruction learning prompting strategy~\citep{ye2023investigating}. 

\textsc{CEBaB}
\begin{prompt}
In a dataset of restaurant reviews there are 4 possible concepts: Good Food,
Good Ambiance, Good Service and Good Noise. Given a certain review, you have
to detect if those concepts are present or not in the review.

Answer format: Good Food:score,Good Ambiance:score, Good Service:score,
Good Noise:score. 

Do not add any text other than that specified by the answer format. 
The score should be equal to 1 if the concept is present or zero otherwise,
no other values are accepted.

The following are examples:

Review: "The food was delicious and the service fantastic".
Answer: Good Food:1, Good Ambiance:0, Good Service:1, Good Noise:0

Review: "The staff was very rough but the restaurant decorations were great.
Other than that there was a very 
relaxing background music".
Answer: Good Food:0, Good Ambiance:1, Good Service:0, Good Noise:1

Now it's your turn: 

Review: <review>       
Answer:
\end{prompt}

\textsc{Drug}
\begin{prompt}
In a dataset of drug reviews there are 2 possible concepts: 

- Effectiveness: 1 if the drug was highly effective and 0 if it was
marginally or not effective,
- Side effects: 1 if the drug gave side effects and 0 otherwise.

Given a certain review, you have to detect if those concepts are present or
not in the review. 

Answer format: Effectveness:score, Side effects:score. 

Do not add any text other than that specified by the answer format. 
The score should be equal to 1 if the concept is present or zero otherwise,
no other values are accepted.

The following are examples:

Review: "The medicine worked wonders for me. However, I did experience some
side effects. Despite this, I still found it easy to use and incredibly
effective".
Answer: Effectiveness:1, Side effects:1

Review: "Not only it did fail to alleviate my symptoms, but it also led to
unpleasant side effects".
Answer: Effectiveness:0, Side effects:1

Now it's your turn:

Review: <review>
Answer:
\end{prompt}

\textsc{Multiemo-it}
\begin{prompt}
In a dataset containing comments in Italian, you need to identify the
following concepts:

-Joy: the user who wrote the comment expresses joy,
-Trust: the user who wrote the comment expresses trust,
-Sadness: the user who wrote the comment expresses sadness,
-Surprise: the user who wrote the comment is surprised.

Response format: Joy:score, Trust:score, Sadness:score, Surprise:score.

The score must be equal to 1 if the concept is present and 0 otherwise; 
other values are not accepted.

The following is an example:
Comment: "Mi piace la rivisitazione di questa canzone, dolce, raffinata,
elegante, bellissima!"
Answer: Joy:1, Trust:1, Sadness:0, Surprise:1

Now it's your turn:
Comment: <comment>
Answer:
\end{prompt}

\textsc{Depression}
\begin{prompt}
You have to identify the presence or absence of 6 concepts in a given text. 
The concepts to be identified are:

- Self-Deprecation: the text exhibits self-critical or self-deprecating
language, expressing feelings
of guilt, shame, or inadequacy. 
- Loss of Interest: diminished pleasure or motivation in the writer's
descriptions of hobbies or pursuits.
- Hopelessness: the writer express feelings of futility or a lack of
optimism about their prospects.
- Sleep Disturbances: the writer mentions insomnia, oversleeping, or
disrupted sleep as part of their 
experience.
- Appetite Changes: there are references to changes in eating habits.
- Fatigue: there are references to exhaustion or lethargy.

Answer format: Self-Deprecation:score, Loss of Interest:score,
Hopelessness:score, Sleep Disturbances:score, Appetite Changes:score,
Fatigue:score.

The score has to be 1 if the concept is detected and 0 otherwise. Do not add
any other text besides the one specified in the answer format.

Text: <text>
Answer:
\end{prompt}

\textsc{IMDb}
\begin{prompt}
In a dataset of film reviews (IMDb), there are 4 possible concepts: 

- Good Acting,
- Good Storyline,
- Good Emotional Arousal,
- Good Cinematography.

Given a certain review, you have to detect if those concepts are 
present or not in the review.

Answer format: 
Good Acting: score, Good Storyline: score, Good Emotional Arousal: score, 
Good Cinematography: score. 

Do not add any text other than that specified by the answer format. 
The score should be equal to 1 if the concept is present and zero if , 
no other values are accepted.

The following are examples:

Review: "The performances were outstanding, especially the lead actor.
The story dragged in the middle though."
Answer: Good Acting: 1, Good Storyline: 0, Good Emotional Arousal: 0, 
Good Cinematography: 0

Review: "This film moved me to tears. The plot was very touching, 
and the visual effects were just stunning."
Answer: Good Acting: 0, Good Storyline: 1, Good Emotional Arousal: 1, 
Good Cinematography: 1

Now it's your turn:

Review: <review>       
Answer:
\end{prompt}

\textsc{TREC-50}
\begin{prompt}
In a dataset of questions, there are 6 possible concepts: 

- Definition Request,
- Person Entity,
- Location Reference,
- Numeric Expectation,
- Abbreviation or Acronym,
- Object Reference.

Given a certain question, you have to detect if those concepts 
are present or not in the question.

Answer format: 
Definition Request: score, Person Entity: score, 
Location Reference: score, Numeric Expectation: score, 
Abbreviation or Acronym: score, Object Reference: score.

Do not add any text other than that specified by the answer format. 
The score should be equal to 1 if the concept is present 
or zero otherwise, no other values are accepted.

The following are examples:

Question: "What is the capital of France?"
Answer: Definition Request: 0, Person Entity: 0, 
Location Reference: 1, Numeric Expectation: 0, 
Abbreviation or Acronym: 0, Object Reference: 0

Question: "Who discovered penicillin?"
Answer: Definition Request: 0, Person Entity: 1, 
Location Reference: 0, Numeric Expectation: 0, 
Abbreviation or Acronym: 0, Object Reference: 0

Now it's your turn:

Question: <review>
Answer:
\end{prompt}

\textsc{CLINC-OOS}
\begin{prompt}
You are given a user query to a task-oriented dialog system. 
The system supports multiple domains and intents, 
but some queries may be out-of-scope (OOS), 
meaning they do not fall into any supported intent.

Your task is to detect the presence or absence of the following 
concepts in the query. For each concept, 
answer with a score of 1 if the concept is present, or 0 if it is absent. 
Do not add any text other than the answer format.

Concepts:
- Domain Mention: Does the query explicitly mention 
or imply a supported domain or topic?
- Intent Specific Keywords: Does the query contain keywords 
or phrases related to any specific intent?
- Action Request: Does the query ask to perform an action or service?
- Out-of-Scope Indicators: Does the query contain terms 
or topics unrelated to any supported domain or intent, 
indicating it is out-of-scope?

Answer format:
Domain Mention: score, Intent Specific Keywords: score, 
Action Request: score, Out-of-Scope Indicators: score

Examples:

Query: "Can you help me book a flight to New York?"
Answer: Domain Mention: 1, Intent Specific Keywords: 1, 
Action Request: 1, Out-of-Scope Indicators: 0

Query: "What's the capital of France?"
Answer: Domain Mention: 0, Intent Specific Keywords: 0, 
Action Request: 0, Out-of-Scope Indicators: 1

Now it's your turn:

Query: <review>
Answer:   
\end{prompt}

\textsc{Banking-77}
\begin{prompt}
In a dataset of user queries related to banking and financial services, 
there are 4 possible concepts:

- Transaction Mention
- Issue/Problem Description
- Account Reference
- Request for Help or Clarification

Given a user query, you have to detect if each of these concepts 
is present or not in the query.

Answer format:  
Transaction Mention: score, Issue/Problem Description: score, 
Account Reference: score, Request for Help or Clarification: score.

Do not add any text other than that specified by the answer format.  
The score should be 1 if the concept is present or 0 otherwise. 
No other values are accepted.

The following are examples:

Query: "A card payment on my account is shown as pending."  
Answer: Transaction Mention: 1, Issue/Problem Description: 1, 
Account Reference: 1, Request for Help or Clarification: 0

Query: "I can't seem to make a standard bank transfer. 
I have tried at least five times already but none of them are going through. 
Please tell me what is wrong?"  
Answer: Transaction Mention: 1, Issue/Problem Description: 1, 
Account Reference: 0, Request for Help or Clarification: 1

Now it's your turn:

Query: <review>  
Answer:
\end{prompt}

\subsection{Experimental Details}
\label{app:exp_details}

For the E2E, CBMs, CEM, DCR and LICEM models, the training process involved utilizing an AdamW optimizer~\citep{adamw}. The $\lambda_y$ coefficient (\ref{eq:loss}) was set to 0.5 to emphasize concept learning over task loss while %$\lambda_w=\num{1e-6}$ 
$\lambda_w=10^{-6}$ and $\lambda_b=10^{-6}$. Moreover, a scheduler was implemented with a gamma of 0.1 and a step size of 10 epochs was employed during the training process, spanning 100 epochs when using BERT as the backbone and 50 epochs when utilizing Mixtral 8x7B. %, with the only exception of LICEM on CEBaB which has been trained for a total of 20 epochs.
After every hidden layer we have used a ReLU activation function.
Here are further insights into the methodologies' architectures, with the number of output neurons indicated within brackets.

\begin{itemize}
  \item E2E: layer 1 (100), layer 2 (number of classes);
  \item CEM: concept embedding size of 768, layer 1 (10), layer 2 (number of classes);
  \item CBMs, concept prediction: layer 1 (10), layer 2 (number of concepts);
      \begin{itemize}
        \item LL, task prediciton: layer (number of classes); 
        \item MLP, task prediction: layer 1 ($3\cdot \text{number of concepts}$), layer 2 (number of classes).
        \end{itemize}
  \item DCR: the temperature parameter is set to $0.1$.  
\end{itemize}

The text's embedding size varies depending on the chosen backbone. When employing BERT, it remains at 768, whereas adopting the LLM approach~\citep{jiang2023scaling} it increases to 4096. For Dtree and XGBoost, we employed the default hyperparameter settings. The DTree model was implemented using the sklearn library, while the XGBoost model was implemented using the xgboost library\footnote{The xgboost library we used can be found at \url{https://github.com/dmlc/xgboost}.}. We conducted five experiments for each methodology. The training time for the different experiments averages around 10 minutes using the setup specified in Section~\ref{sec:setup}.

%Regarding the concept annotated datasets, the CEBaB dataset~\citep{abraham2022cebab} does not necessitate any splitting procedure as it inherently offers training, validation, and test sets. In the training set, modifications include counterfactual examples, while both the validation and test sets exclusively contain original reviews. For the remaining datasets, we partitioned the data into training, validation, and test sets using stratified sampling based on the task labels. The proportions allocated are 0.7 for training, 0.1 for validation, and 0.2 for testing. 

Since concept-annotated datasets were discussed in Section~\ref{sec:setup}, we focus here on the remaining datasets. \textbf{Depression} contains Reddit posts from users and the goal is to classify a post as depressed or not. As concept annotations are unavailable, we used an LLM~\citep{jiang2024mixtral} to identify six relevant concepts: \textit{Self-deprecation}, \textit{Loss of Interest}, \textit{Hopelessness}, \textit{Sleep Disturbances}, \textit{Appetite Changes}, and \textit{Fatigue}. \textbf{IMDb} includes 50,000 movie reviews labeled as positive or negative for sentiment analysis. The LLM identified: \textit{Acting}, \textit{Storyline}, \textit{Emotional Arousal}, and \textit{Cinematography}. \textbf{TREC-50} comprises open-domain questions classified into 50 fine-grained types. Relevant concepts identified: \textit{Definition Request}, \textit{Person Entity}, \textit{Location Reference}, \textit{Numeric Expectation}, \textit{Abbreviation or Acronym}, and \textit{Object Reference}. \textbf{Banking-77} features real-world banking queries labeled with 77 intent categories. The LLM-generated concepts are: \textit{Transaction Mention}, \textit{Issue/Problem Description}, \textit{Account Reference}, and \textit{Request for Help or Clarification}. \textbf{CLINC-OOS} contains 151 in-domain intents across ten topics and one out-of-scope class, used for open-domain intent classification. The associated concepts are \textit{Domain Mention}, \textit{Intent-Specific Keywords}, \textit{Action Request}, and \textit{Out-of-Scope Indicators}. All datasets included predefined training and test splits. Additionally, \textbf{IMDb} and \textbf{CEBaB} provided validation sets. For the remaining datasets, we split the training such that $\frac{1}{8}$ of the data is used as validation.

\subsection{Encoder comparison}
\label{app:enc_comp}
This section presents all the results obtained using a fine-tuned BERT backbone as the encoder $h(x)$. In the remainder of the paper, we consistently reported results when utilizing Mixtral 8x7B~\citep{jiang2024mixtral} as the backbone model. The total number of trainable parameters remains relatively modest: approximately {100K} when using Mixtral as the backbone (without fine-tuning), and around {10M} when using BERT as the backbone.
In this section, we provide the performance of all models in terms of task accuracy (see Table~\ref{tab:task_acc_bert}) and of concept macro-averaged F1 score (refer to Table~\ref{tab:conc_acc_bert}) when employing BERT as the backbone~\citep{devlin2018bert}, which is an encoder-only model. 

\begin{table}[h!]
\centering
\caption{This table presents the performance in terms of task accuracy (\%) of different models utilizing BERT as backbone. We report in \textbf{bold} the best result among the same type of models (e.g., supervised, interpretable ones) considering models equally best if their standard deviations overlap. We use \cmark  to indicate models requiring concept supervision (C. Sup.) or having an interpretable task predictor (T. Inter.). We highlight in light gray the models we propose in this work. We do not report supervised model results for depression ($-$) since it does not provide concept annotations.}
\label{tab:task_acc_bert}
\setlength{\tabcolsep}{4pt}
% \resizebox{\textwidth}{!}{
{\small
\begin{tabular}{l|lcc|cccc}
\toprule
Type & Method        & C. S.   & T. I.  & CEBaB           & Multiemo-It        & Drug   & Depression\\
\midrule
\textsc{e2e} 
& MLP         & \xmark & \xmark  & \textbf{90.68}\tiny{ $\pm$ 0.47}  & \textbf{75.67}\tiny{$\pm$ 0.47}   & \textbf{59.33}\tiny{ $\pm$ 0.56}  &  \textbf{97.80}\tiny{ $\pm$0.23}  \\
\midrule
\multirow{7}{*}{\textsc{sup.}} %{\rotatebox[origin=c]{90}{\textsc{supervised}}}
& CBM+MLP      & \cmark & \xmark & 78.01\tiny{ $\pm$ 6.51}   & 54.10\tiny{ $\pm$ 4.51}   & 36.67\tiny{ $\pm$ 6.24}  &--\\ 
& CBM+XG       & \cmark & \xmark & 80.00 \tiny{ $\pm$ 0.34}   & 69.02\tiny{ $\pm$ 0.64}   & 51.00\tiny{ $\pm$ 0.28}  &--\\ 
& CEM          & \cmark & \xmark & \textbf{90.67}\tiny{ $\pm$ 0.47}   & \textbf{77.00}\tiny{ $\pm$ 0.82}   & \textbf{58.33}\tiny{ $\pm$ 1.70}  &--\\ 
\cmidrule{2-8}
& CBM+LL       & \cmark & \cmark  & 61.00\tiny{ $\pm$ 12.02}   & 49.67\tiny{ $\pm$ 5.46}   & 34.33\tiny{ $\pm$ 7.38} &--\\ 
& CBM+DT       & \cmark & \cmark  & 75.67\tiny{ $\pm$ 0.47}   & 65.02\tiny{ $\pm$ 0.34}   & 46.23\tiny{ $\pm$ 0.78}  &--\\ 
& DCR       & \cmark & \cmark  & 86.55\tiny{ $\pm$ 0.58}   & 74.01\tiny{ $\pm$ 0.24}   & \textbf{59.75}\tiny{ $\pm$ 0.45}  &--\\ 
& \cellcolor{customgray!20}LICEM (ours)      & \cellcolor{customgray!20}\cmark & \cellcolor{customgray!20}\cmark  & \cellcolor{customgray!20}\textbf{87.89}\tiny{ $\pm$ 0.38}   & \cellcolor{customgray!20}\textbf{75.31}\tiny{ $\pm$ 0.15}   & \cellcolor{customgray!20}\textbf{60.14}\tiny{ $\pm$ 0.44}  &\cellcolor{customgray!20}-- \\ 
\midrule
\multirow{7}{*}{\textsc{gen.}}%\rotatebox[origin=c]{90}{\textsc{generative}}}
& CBM+MLP      & \xmark & \xmark & 73.93\tiny{ $\pm$ 5.67}   & 44.19\tiny{ $\pm$ 2.07}   & 35.16\tiny{ $\pm$ 4.3} & 83.20\tiny{ $\pm$ 2.18}  \\ 
& CBM+XG       & \xmark & \xmark & 83.29\tiny{ $\pm$ 0.43}   & 69.85\tiny{ $\pm$ 1.55}   & 34.94\tiny{ $\pm$ 0.91} & 87.00\tiny{ $\pm$ 1.01}  \\ 
& CEM          & \xmark & \xmark & \textbf{85.88}\tiny{ $\pm$ 0.95}   & \textbf{73.15}\tiny{ $\pm$ 0.67}   & \textbf{56.95}\tiny{ $\pm$ 0.36} &   \textbf{96.12}\tiny{ $\pm$ 0.50}  \\ 
\cmidrule{2-8}
& CBM+LL       & \xmark & \cmark  & 58.81\tiny{ $\pm$ 7.16}   & 58.35\tiny{ $\pm$ 1.59}   & 36.84\tiny{ $\pm$ 11.52} &   51.48\tiny{ $\pm$ 2.16}  \\ 
& CBM+DT       & \xmark & \cmark  & 79.28\tiny{ $\pm$ 0.52}   & 62.61\tiny{ $\pm$ 2.08}   & 34.17\tiny{ $\pm$ 0.11} &   80.55\tiny{ $\pm$ 0.03}  \\ 
& DCR       & \xmark & \cmark  & \textbf{85.63}\tiny{ $\pm$ 0.81}   & 70.02\tiny{ $\pm$ 2.70}   & 57.46\tiny{ $\pm$ 0.02} &   95.98\tiny{ $\pm$ 0.27}  \\ 
& \cellcolor{customgray!20}LICEM (ours)  & \cellcolor{customgray!20}\xmark & \cellcolor{customgray!20}\cmark  & \cellcolor{customgray!20}\textbf{86.22}\tiny{ $\pm$ 0.66}   & \cellcolor{customgray!20}\textbf{74.45}\tiny{ $\pm$ 0.57}   & \cellcolor{customgray!20}\textbf{60.23}\tiny{ $\pm$ 0.58} &   \cellcolor{customgray!20}\textbf{96.87}\tiny{ $\pm$ 0.20}  \\ 
% \midrule
% \multirow{2}{*}{\begin{tabular}{@{}l@{}}\textsc{self}\\\textsc{sup.}\end{tabular}} 
% & R-ICEM  & \xmark & \cmark & 86.7\tiny{ $\pm$ 0.25}   & 80.47\tiny{ $\pm$ 0.63}   & 60.38\tiny{ $\pm$ 0.84} & 95.44\tiny{$\pm$ 0.11} \\ 
% & L-ICEM  & \xmark & \cmark & \textbf{87.03}\tiny{ $\pm$0.16}   & \textbf{82.71$^*$}\tiny{ $\pm$ 0.42}   & \textbf{61.25$^*$}\tiny{ $\pm$ 0.75} & \textbf{97.03$^*$}\tiny{ $\pm$ 0.65} \\ 
% \midrule
% \multirow{2}{*}{\begin{tabular}{@{}l@{}}\textsc{self}$^2$\\\textsc{sup.}\end{tabular}} & R-ICEM & \xmark & \xmark &     &         &   & \\
% & L-ICEM\\
\bottomrule
\end{tabular}
}
\end{table}

\begin{table}[h!]%{R}{9.cm}
\centering
\caption{This table presents the performance in terms of concept prediction of the models that utilize BERT as backbone. Concept prediction (\%) of the compared models for datasets equipped with concept annotations is measured using the macro-averaged F1 score. We report in \textbf{bold} the best result among the same type of models (e.g., supervised, interpretable ones) considering models equally best if their standard deviations overlap. We highlight in light gray the models we propose in this work. The methods using the self-generative have the same macro-averaged F1 score, therefore we use $-$ to represent all methods.}
\label{tab:conc_acc_bert}
% \resizebox{9cm}{!}{
\begin{tabular}{l|l|cccc}
\toprule
Type & Method        & CEBaB                               & Multiemo-It        &               Drug    \\
\midrule
\textsc{e2e}
& MLP            & \textbf{79.92}\tiny{$\pm$ 1.77}                      & \textbf{63.25}\tiny{ $\pm$ 1.09}        &  \textbf{79.01}\tiny{ $\pm$ 2.9}  \\
\midrule
\multirow{7}{*}{\textsc{sup.}} %{\rotatebox[origin=c]{90}{\textsc{supervised}}}
& CBM+MLP       & 75.17\tiny{$\pm$ 3.11}                      & \textbf{64.08}\tiny{$\pm$ 1.22}        & 74.26\tiny{$\pm$ 0.9} \\ 
& CEM           & \textbf{79.97}\tiny{$\pm$ 1.29}                      & \textbf{64.42}\tiny{$\pm$ 1.21}        & \textbf{77.32}\tiny{$\pm$ 1.2} \\ 
& CBM+XG           & \textbf{79.92}\tiny{$\pm$ 1.77}                      & \textbf{63.25}\tiny{ $\pm$ 1.09}        & \textbf{79.01}\tiny{ $\pm$ 0.9} \\ 

\cmidrule{2-5}
& CBM+LL        & 74.25\tiny{$\pm$ 4.55}                      & 62.08\tiny{$\pm$ 0.88}       & 73.11\tiny{$\pm$ 1.7} \\ 
& CBM+DT        & 79.92\tiny{$\pm$ 1.77}                     & 63.25\tiny{ $\pm$ 1.09}       & 79.01\tiny{ $\pm$ 2.9} \\ 
& DCR       & 82.06\tiny{$\pm$ 0.40}                    & 64.29\tiny{$\pm$ 0.42}              & 80.10\tiny{$\pm$ 0.2}       \\ 
& \cellcolor{customgray!20}LICEM (ours)        & \cellcolor{customgray!20} \textbf{82.93}\tiny{ $\pm$ 0.13}   & \cellcolor{customgray!20} \textbf{65.61}\tiny{ $\pm$ 0.69} & \cellcolor{customgray!20} \textbf{81.59}\tiny{ $\pm$ 0.42} \\ 
\midrule
\multirow{7}{*}{\textsc{gen.}}%\rotatebox[origin=c]{90}{\textsc{generative}}} 
& CBM+MLP       & 75.05\tiny{ $\pm$ 8.31}                      & 49.59\tiny{ $\pm$ 10.01}        & 43.58\tiny{ $\pm$ 14.99} \\ 
& CEM           & \textbf{81.08}\tiny{ $\pm$ 0.44}                      & \textbf{58.30}\tiny{ $\pm$ 1.79}        & \textbf{80.99}\tiny{ $\pm$ 0.42} \\
& CBM+XG                & 79.24\tiny{ $\pm$ 1.21}                      & \textbf{60.79}\tiny{ $\pm$ 0.71}        & 64.72\tiny{ $\pm$ 0.45} \\ 
\cmidrule{2-5}
& CBM+LL        & \textbf{78.75}\tiny{ $\pm$ 0.59}                    & \textbf{61.72}\tiny{ $\pm$ 0.24}         & 66.72\tiny{ $\pm$ 19.48}  \\
& CBM+DT           & \textbf{79.24}\tiny{ $\pm$ 1.21}                      & \textbf{60.79}\tiny{ $\pm$ 0.70}        & 64.72\tiny{ $\pm$ 0.45} \\ 
& DCR       & \textbf{80.25}\tiny{ $\pm$ 1.02}                    & 59.11\tiny{ $\pm$ 0.84}         & \textbf{81.47}\tiny{ $\pm$ 0.49} \\ 
& \cellcolor{customgray!20}LICEM (ours)     & \cellcolor{customgray!20}\textbf{77.79}\tiny{ $\pm$ 2.49}     & \cellcolor{customgray!20}58.87\tiny{ $\pm$ 0.66}         & \cellcolor{customgray!20}\textbf{81.18}\tiny{ $\pm$ 0.33}\\ 
\midrule
\multirow{1}{*}{\begin{tabular}{@{}l@{}}\textsc{self gen.}\end{tabular}} 
& \cellcolor{customgray!20} --   & \cellcolor{customgray!20} \textbf{84.08}\tiny{ $\pm$ 0.00}                    & \cellcolor{customgray!20} \textbf{64.27}\tiny{ $\pm$0.00}          & \cellcolor{customgray!20} \textbf{83.00}\tiny{ $\pm$0.00} \\ 
\bottomrule
\end{tabular}
% }
\end{table}

Both tables show that there is no great difference with respect to Tables~\ref{tab:task_acc},~\ref{tab:conc_acc}, with BERT providing slightly lower performance on Multiemo-It and on the Drug dataset. This result shows that the proposed approach can be applied also to other architectures. We chose to employ Mixtral in the remainder of the paper since it can be also effectively used to provide concept annotations (thus enabling the Self-LICEM strategy), therefore having a single model for both encoding the sample and predicting the concept predictions.

\subsection{LLM-based concept annotation vs Class-level annotation}
\label{app:annotation_comparison}
\begin{figure}[t]
    \centering
    \includegraphics[width=0.6\textwidth]{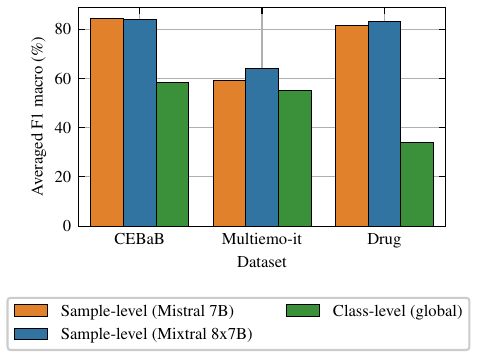}
    \caption{Comparison among concept annotation methods where the annotation quality is measured in terms of macro-averaged F1 score. On average, \hbox{Mixtral 8x7B} yields the best results.}
    \label{fig:global_local}
\end{figure}

This section presents a comparison between the usage of two different LLMs, Mistral 7B~\citep{jiang2023mistral} and Mixtral 8x7B~\citep{jiang2024mixtral}, as concept annotators. In Fig.~\ref{fig:global_local} we report the results in terms of macro-averaged F1 score (as
concept classes are highly imbalanced) on the three datasets for which human concept annotation is available. We also report, as a baseline, a global (class-level) annotation strategy, providing to all samples belonging to a given class the same concept annotation. 
In this case, we label the positive class with positive concepts and negated negative concepts (e.g. for all samples of the class \textit{Good Drug} we use `Efficient' and 'Not Side Effects'). We can observe that between the two LLMs there is not a significant difference in performance, with \hbox{Mixtral 8x7B} providing on average slightly better results. Comparing against the baseline, instead, we can observe that there is a great improvement in CEBaB and in the Drug dataset, while in Multiemo-It the improvement is more modest. 

\subsection{Concepts prediction performance}
\label{app:conc_acc}
In this section we report in Table~\ref{tab:conc_acc} the averaged F1 macro to measure the concepts prediction performance of all models when provided with all the available concept annotations for all the different experiments conducted, generative approach included, when using \hbox{Mixtral 8x7B} as a backbone. 
%As shown in Table~\ref{tab:task_acc_complete}, LICEM outperforms the other task interpretable models, reaching the highest task accuracy for the CEBaB dataset using the generative approach.
 The results shown in Fig.~\ref{fig:efficiency} are here confirmed. We again see that Self-supervised strategy is a very good approach since without human effort it provides better concept macro-averaged F1 score in CEBaB and Drug. Only on Multiemo-It the performance are significantly lower. This result may be due to the fact that the latter dataset is in Italian while the other datasets are in English, a language for which the LLMs have certainly seen more training samples. 

\input{conc_acc}

\subsection{Concept interventions}
\begin{figure}[h!]
    \centering
    \includegraphics[width=0.8\textwidth]{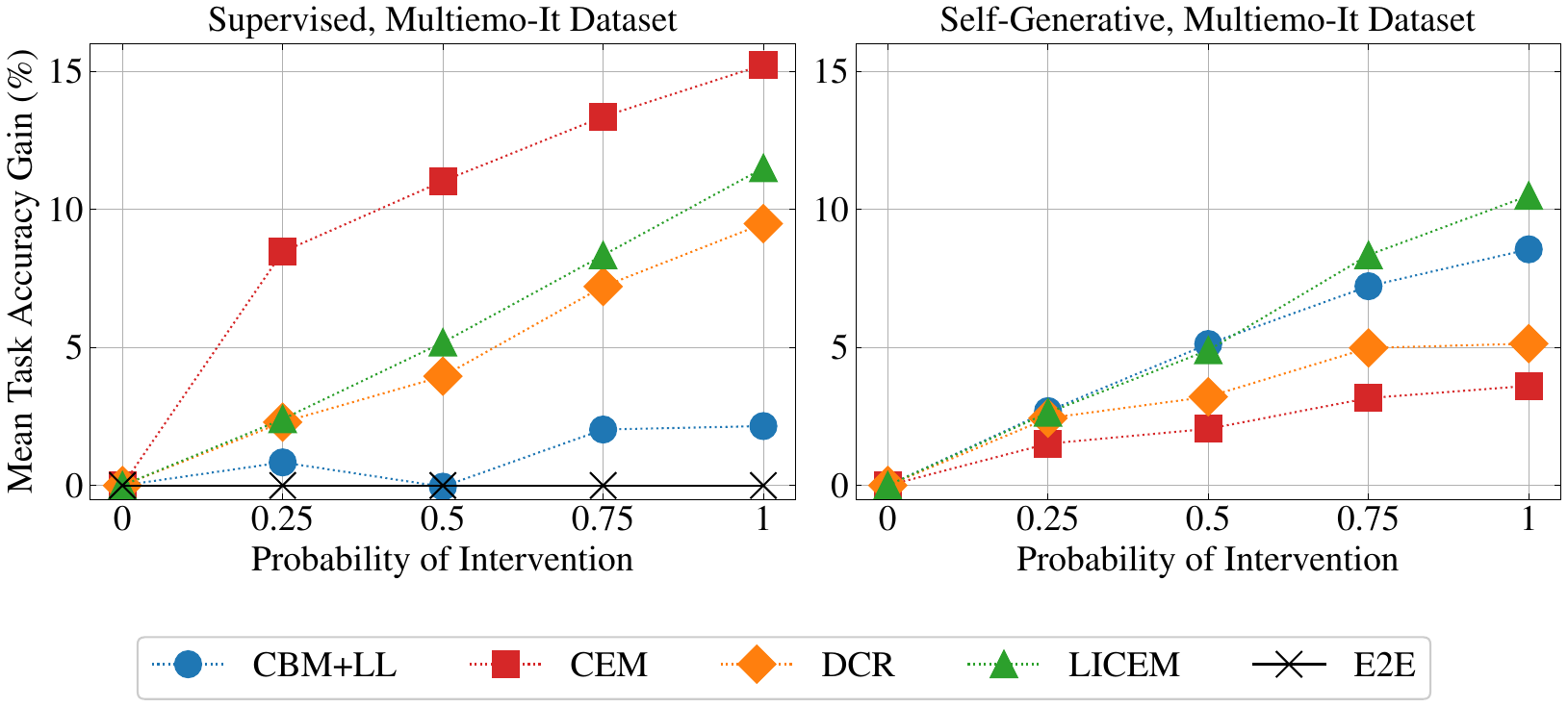}
    \caption{Concept interventions on the Multiemo-it dataset for (left) supervised approaches and (right) self-supervised ones.}
    \label{fig:concept_int_emo}
\end{figure}

\begin{figure}[h!]
    \centering
    \includegraphics[width=0.8\textwidth]{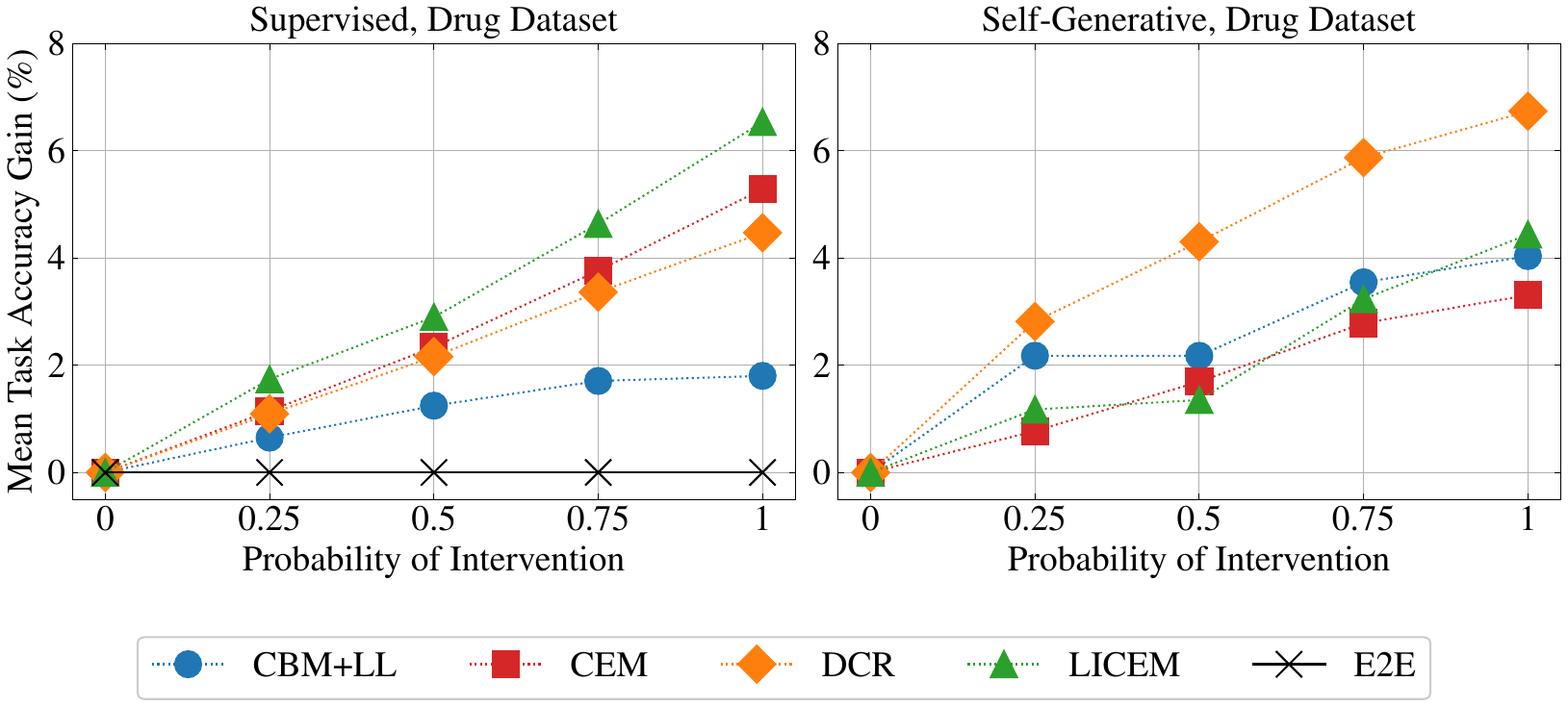}
    \caption{Concept interventions on the Drug dataset for (left) supervised approaches and (right) self-supervised ones.}
    \label{fig:concept_int_drug}
\end{figure}

As introduced in Section~\ref{sec:interpretability}, LICEM is sensible to concept interventions. This characteristic is very important since it implies that a human can interact with the model, providing counterfactual predictions when prompted with different concept predictions. In Fig.~\ref{fig:app_concept_int},~\ref{fig:concept_int_emo},~\ref{fig:concept_int_drug} we simulate this situation by correcting mispredicted concepts with the correct concept predictions and check whether the task prediction has been also modified. More in details, we report the improvement in task accuracy when increasing the probability to correct the concepts, demonstrating LICEM's responsiveness and significant performance improvement. A similar behaviour can also be observed for CBMs, even though they were starting from a lower task accuracy and a higher increase could have also been expected.
For comparison, we also report the E2E model with a flat line, since it does not offer this possibility. As noted in~\citep{zarlenga2022concept}, CEMs (which are not task interpretable) may not respond well to concept interventions, especially without conducting them during training. Thus, we trained all CEM-based models with a 0.5 intervention probability during the forward pass. 

\label{app:conc_int}
\begin{figure}[h!]
    \centering
    \includegraphics[width=0.8\textwidth]{cebab_intervention_plot.pdf}
    \caption{Concept interventions on the CEBaB dataset for (left) supervised approaches and (right) self-supervised ones.}
    \label{fig:app_concept_int}
\end{figure}

% \subsection{Causal Concept Effect (CaCE)}
% \label{app:cace}
% \input{sup_cace}
% %\input{unsup-cace}
% As anticipated in Section~\ref{sec:interpretability}, Concept-based models predictions must be causally influenced by the predicted concepts. We assess concept-based models' responsiveness to \textit{do-interventions} using the Causal Concept Effect (CaCE)~\citep{goyal2019explaining}, which measures the impact of input modifications on model predictions. Higher absolute CaCE values indicate stronger conditioning on relevant concepts. Tables~\ref{tab:cace_sup} shows that both supervised and self-supervised LICEM have higher CaCE values compared to CBM, CEM and DCR, suggesting stronger reliance on predicted concepts. %, while low CaCE values for CBM. 
% This result is positive since all concepts considered in this work are relevant for the task at hand. We leave for future work the exploration of tasks where there are confounding concepts and checking whether LICEM is capable to not consider them. 

\subsection{Additional visualizations of explanations}
\label{app:explanations}
\begin{figure}[h]
    \centering
    \includegraphics[width=0.95\textwidth]{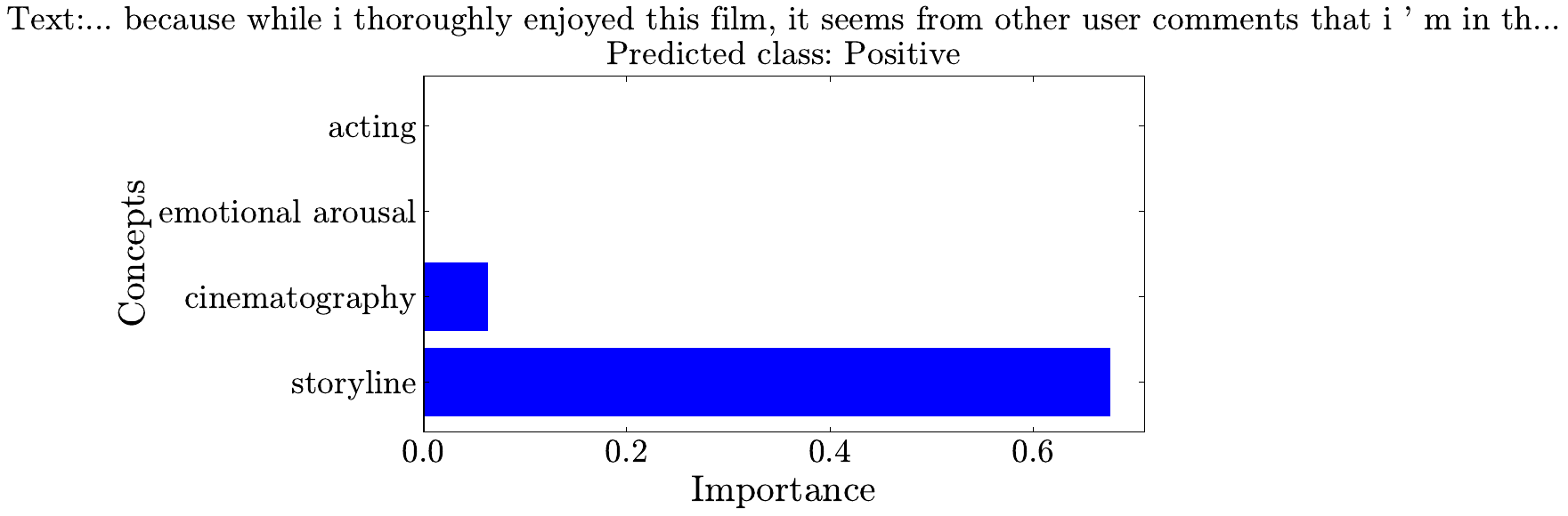}
    \caption{Explanation 1 from the IMDB dataset.}
    \label{fig:imdb_1}
\end{figure}

\begin{figure}[h]
    \centering
    \includegraphics[width=0.95\textwidth]{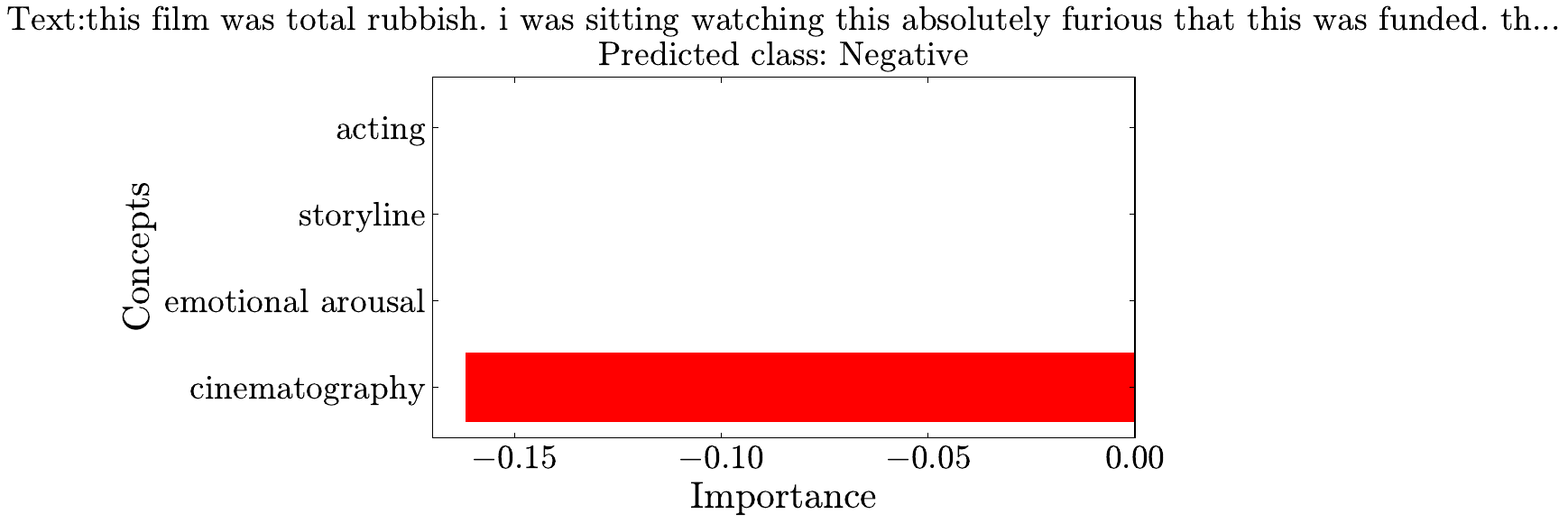}
    \caption{Explanation 2 from the IMDB dataset.}
    \label{fig:imdb_2}
\end{figure}

\begin{figure}[h]
    \centering
    \includegraphics[width=0.95\textwidth]{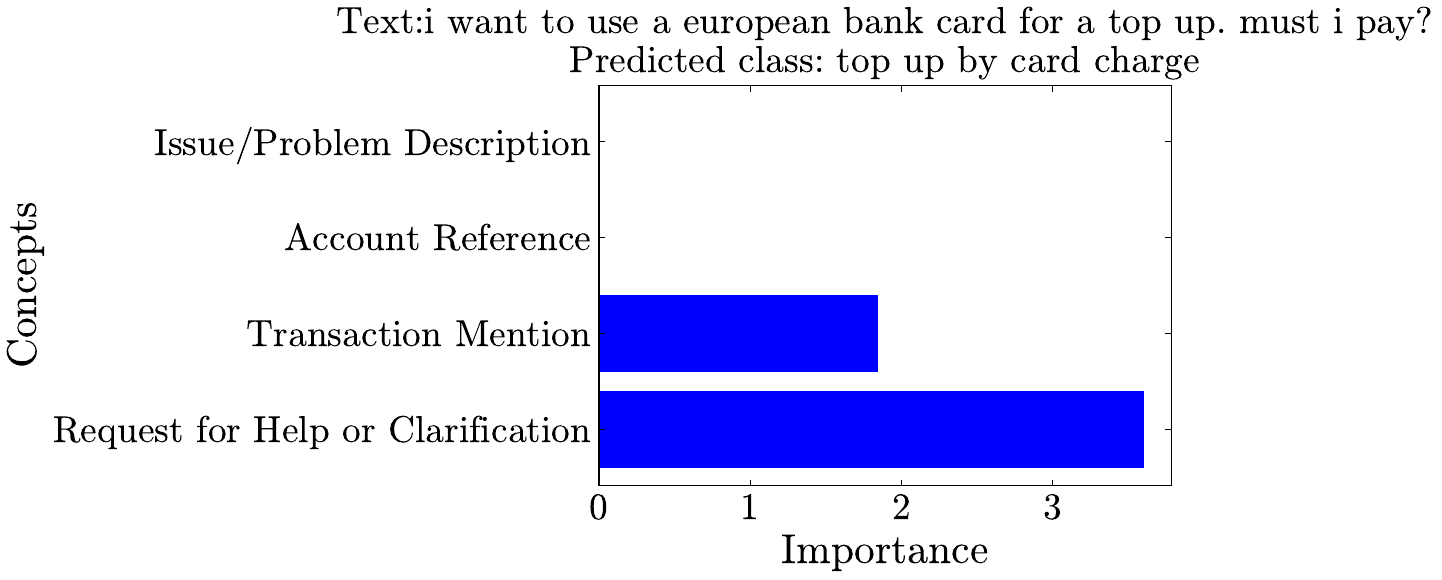}
    \caption{Explanation 1 from the Banking-77 dataset.}
    \label{fig:bank_1}
\end{figure}

\begin{figure}[h]
    \centering
    \includegraphics[width=0.8\textwidth]{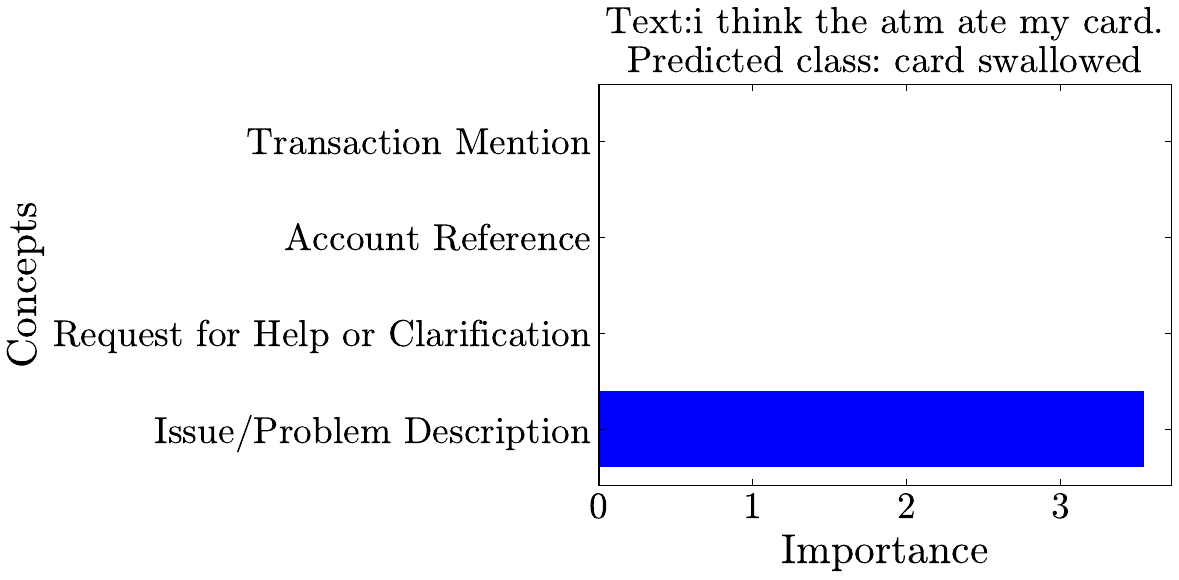}
    \caption{Explanation 2 from the Banking-77 dataset.}
    \label{fig:bank_2}
\end{figure}

\begin{figure}[h]
    \centering
    \includegraphics[width=0.8\textwidth]{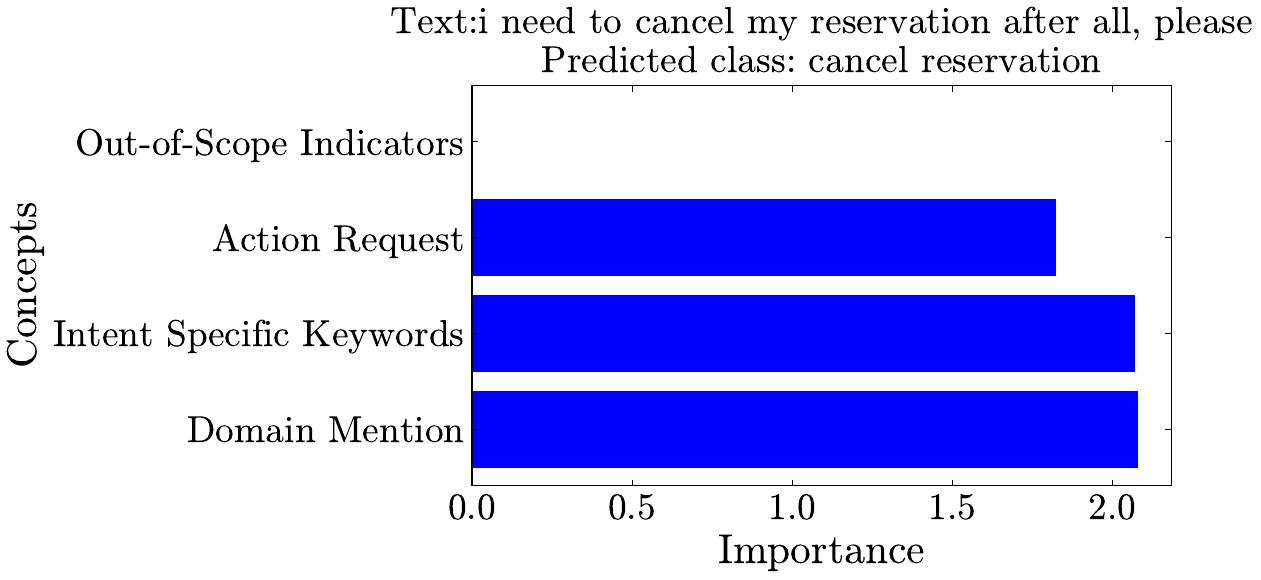}
    \caption{Explanation 1 from the CLINC-OOS dataset.}
    \label{fig:clinc_1}
\end{figure}

\begin{figure}[h]
    \centering
    \includegraphics[width=0.8\textwidth]{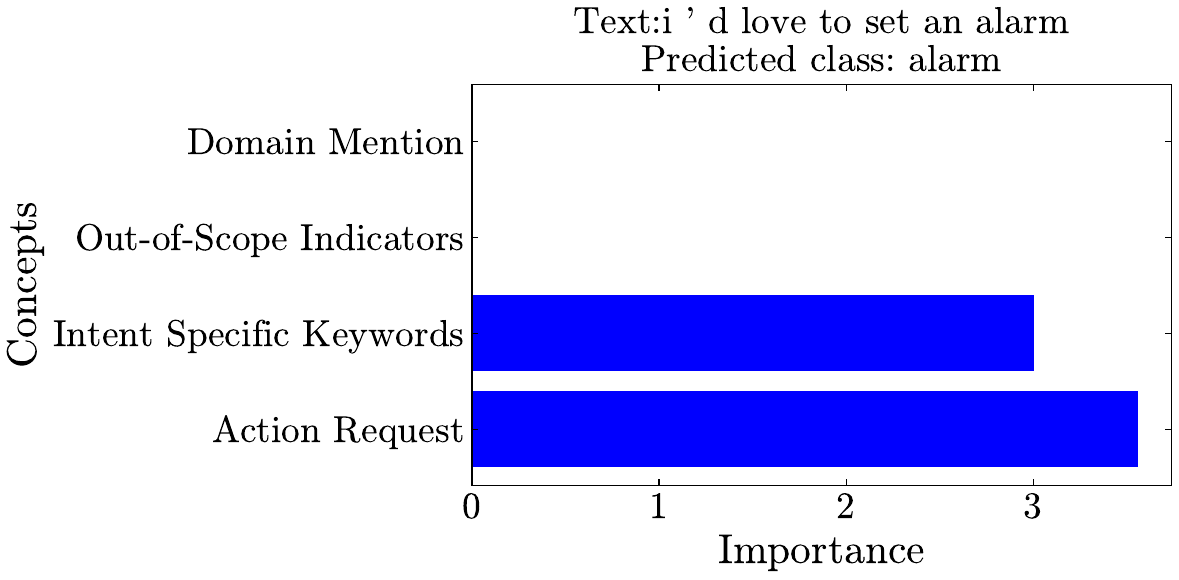}
    \caption{Explanation 2 from the CLINC-OOS dataset.}
    \label{fig:clinc_2}
\end{figure}

\begin{figure}[h]
    \centering
    \includegraphics[width=0.8\textwidth]{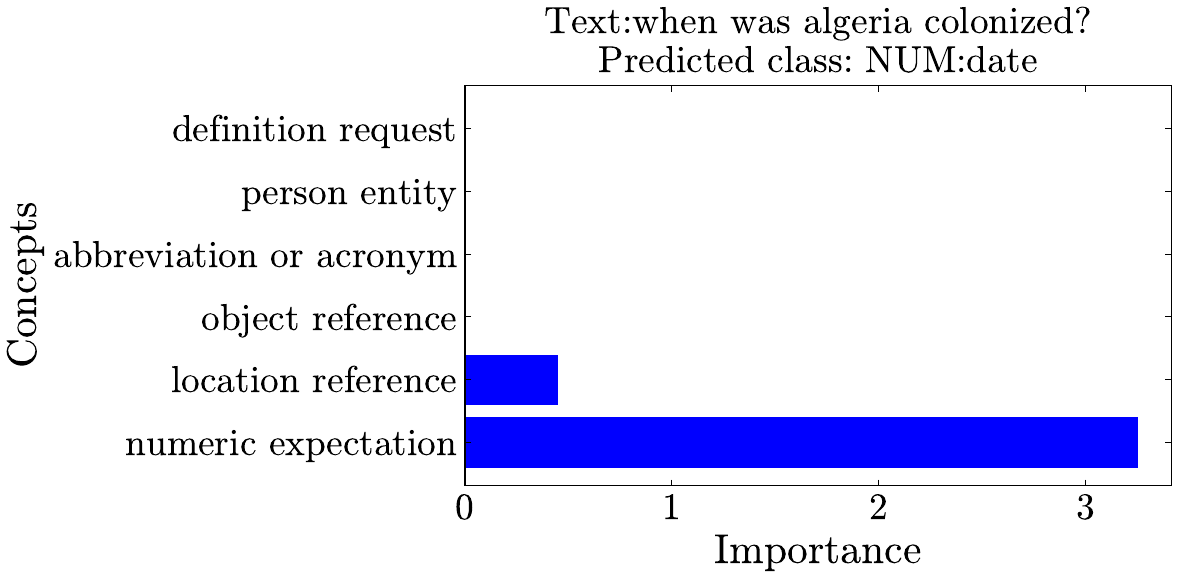}
    \caption{Explanation 1 from the TREC-50 dataset.}
    \label{fig:trec_1}
\end{figure}

\begin{figure}[h]
    \centering
    \includegraphics[width=0.8\textwidth]{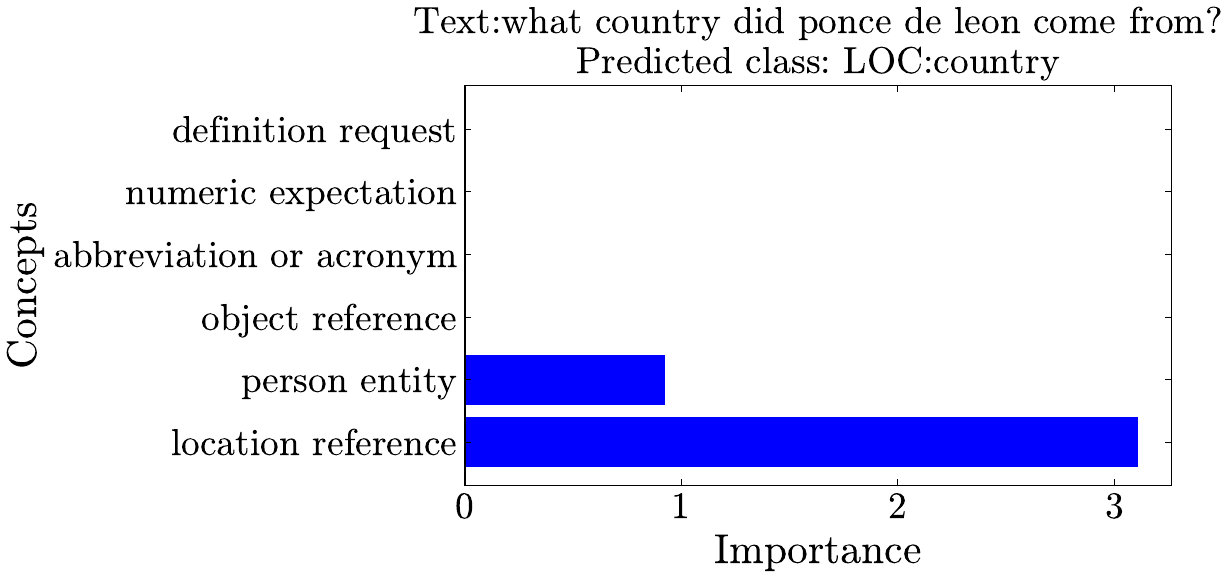}
    \caption{Explanation 2 from the TREC-50 dataset.}
    \label{fig:trec_2}
\end{figure}

This appendix provides additional LICEM explanations. Although the main text includes a representative set of explanations to support the core findings, the materials presented here offer a broader view of the model’s interpretability across different datasets. From Fig.~\ref{fig:imdb_1} to Fig.~\ref{fig:trec_2}, the concepts on the y-axis are ordered in ascending order. This means that the most important concepts—those with the highest importance—are displayed at the bottom of the y-axis, with decreasing importance as you move upward.

\subsection{User-Study Characterization}
\label{app:user_study}

\begin{figure}[h!]
    \centering
    \includegraphics[width=0.8\textwidth]{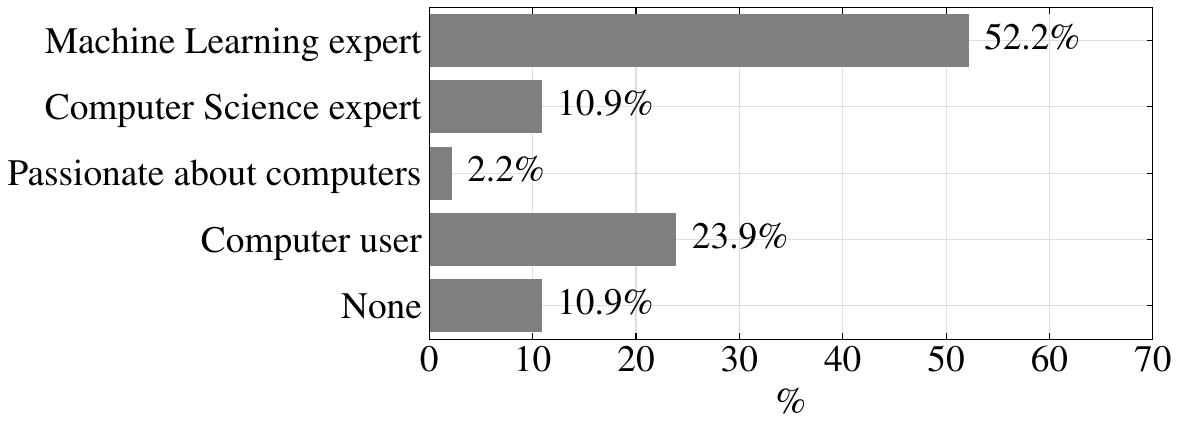}
    \caption{Distribution of users by expertise level.}
    \label{fig:survey_info}
\end{figure}

In this section, we provide further details regarding the conducted survey. A total of $46$ participants with varying levels of experience in machine learning, from complete beginners to experts, were recruited (see Fig.~\ref{fig:survey_info}). The gender distribution was nearly balanced, with $40\%$ identifying as female and $60\%$ as male. The majority of participants, $91.3\%$, were within the $20-40$ age range, while only $8.7\%$ were aged over $40$.

\begin{figure}[h!]
    \centering
    \includegraphics[width=1\textwidth]{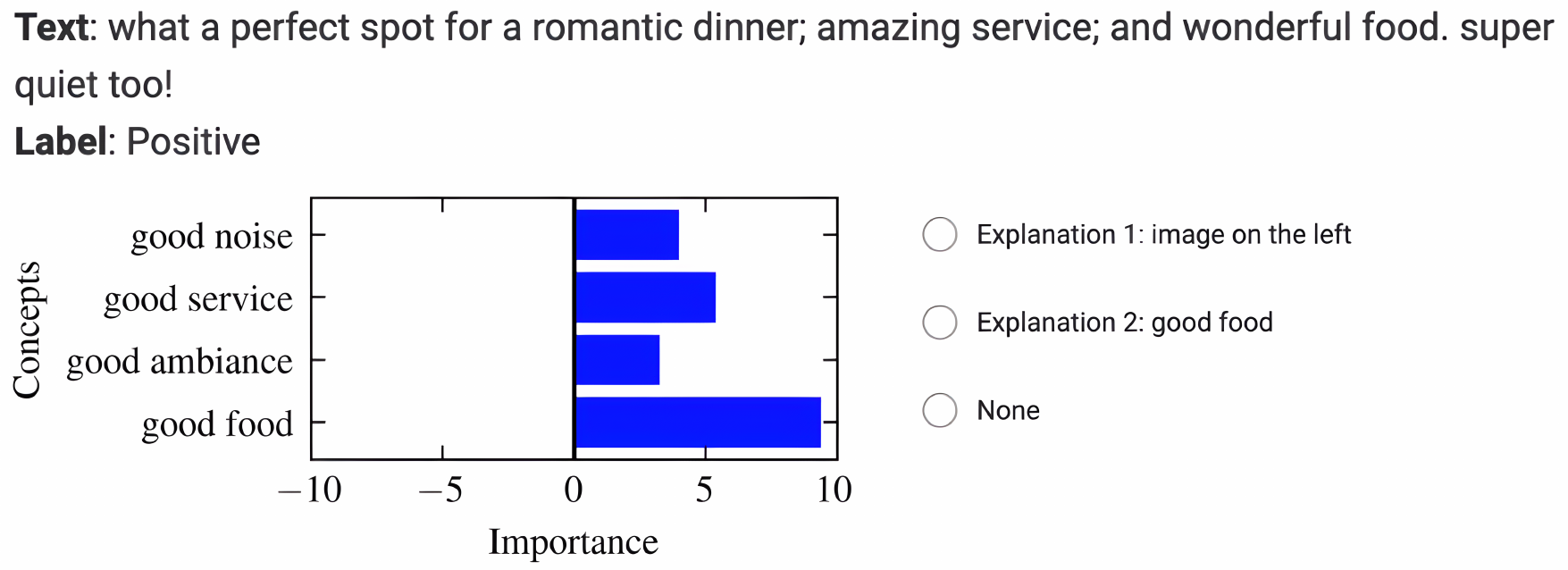}
    \caption{Example of explanation plausibility question, CEBaB dataset.}
    \label{fig:cebab_preference}
\end{figure}

\begin{figure}[h!]
    \centering
    \includegraphics[width=0.8\textwidth]{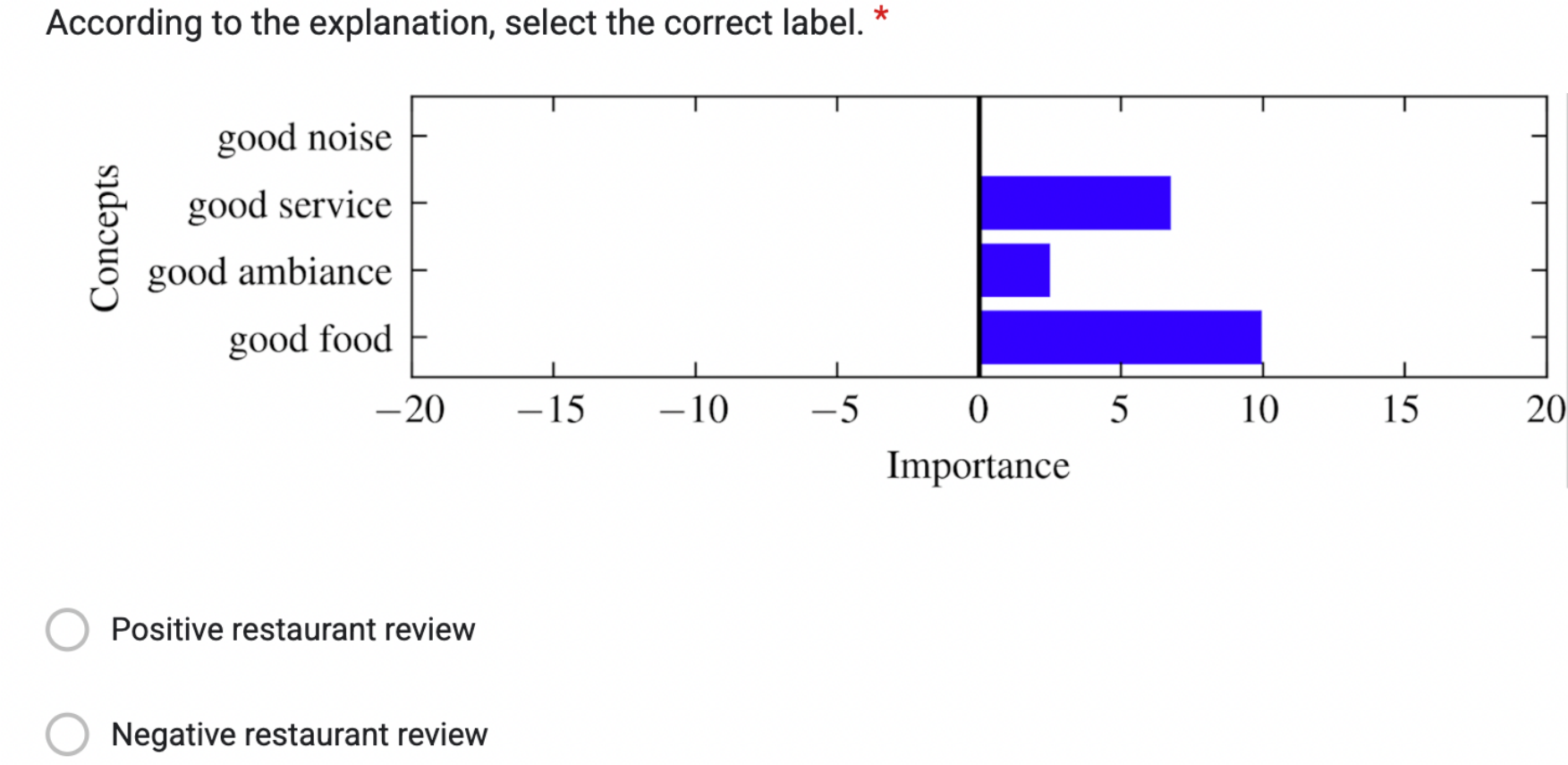}
    \caption{Example of label prediction given LICEM explanation, CEBaB dataset.}
    \label{fig:cebab_prediction_licem}
\end{figure}

%and asked to predict the label of the corresponding samples based on that explanation.   

\begin{comment}
\begin{figure}[h]
    \centering
    \includegraphics[width=1\textwidth]{report_main.pdf} %%%TODO: rimpiazzare
    \caption{Averaged survey results for the two user groups. On the left, we report the explanation plausibility; on the right, users' accuracy in guessing the model prediction based on its explanation.}
    \label{fig:survey}
\end{figure}
\end{comment}

\subsection{Critical Difference Diagram}
\label{app:cd_diagram}

The Critical Difference (CD) diagram was computed by considering all the models across all datasets for both the generative and self-generative approaches. The supervised approach was not included, as the depression dataset was unavailable (no concept supervision). However, even in the supervised scenario, LICEM and CEM remain the top-performing models.

\begin{figure}[h!]
    \centering
    \includegraphics[width=1\textwidth]{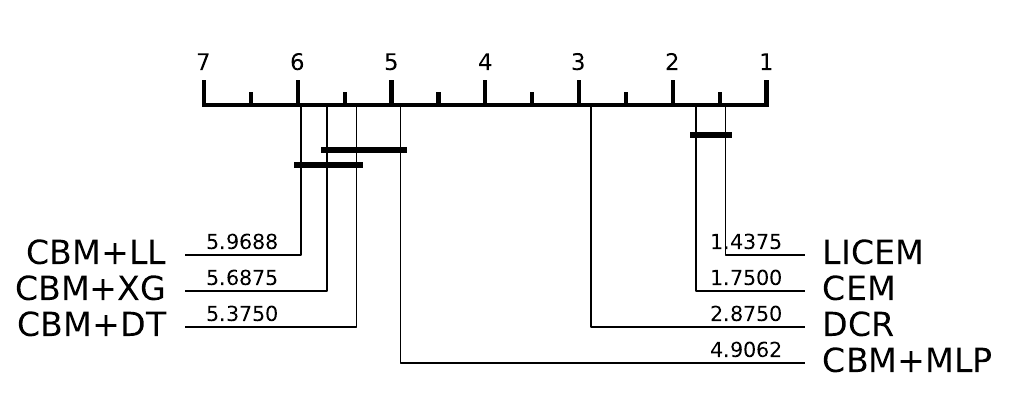}
    \caption{Results for the CD diagram considering all the models in both the \textit{supervised},  \textit{generative}, and \textit{self-generative} scenarios for all the datasets.}
    \label{fig:cd_results}
\end{figure}

The results of the CD diagram (Figure~\ref{fig:cd_results}) indicate that LICEM is the top performer, consistently achieving first or second place across all datasets. However, the performance difference between LICEM and CEM is not statistically significant, nor is there a notable distinction between the CBM-like models (CBM+MLP, CBM+XG, and CBM+LL). DCR ranks in the middle tier with an average position of 3.0, demonstrating strong performance, although it does not outperform LICEM or CEM. Both CBM+XG and CBM+MLP also rank in the middle, while CBM+LL consistently ranks the lowest across all datasets.

\subsection{Concept Sparsity}
\label{app:sparsity}

In this section, we provide a quantitative analysis of how closely the predicted concepts sparsity of each model matches the intrinsic sparsity of the concepts in the data. For each sample, we compute the absolute difference between the number of predicted active concepts and the number of active concepts in the ground truth:
\[
\delta_i = \left|\sum_j \hat{c}_{ij} - \sum_j c_{ij}\right|,
\]
where $\hat{c}_{ij}\in\{0,1\}$ denotes whether concept $j$ was predicted as active in sample $i$, and $c_{ij}\in\{0,1\}$ is the corresponding ground-truth label. The average deviation across all samples provides a direct measure of this alignment, with lower values indicating better correspondence.
Table~\ref{tab:avg_deviation} reports the results across all datasets.

\begin{table}[h!]
\caption{Average deviation between predicted and ground-truth sparsity. Lower values indicate closer alignment with the intrinsic sparsity of the data.}
\label{tab:avg_deviation}
\centering
\begin{tabular}{lccc}
\toprule
\textbf{Method} & \textbf{DRUG} & \textbf{CEBaB} & \textbf{MULTIEMO} \\
\midrule
CBM + LL & 0.48 $\pm$ 0.07 & 0.64 $\pm$ 0.09 & 0.66 $\pm$ 0.05 \\
CBM + MLP & 0.30 $\pm$ 0.03 & 0.58 $\pm$ 0.04 & 0.71 $\pm$ 0.06 \\
CEM & 0.21 $\pm$ 0.02 & 0.42 $\pm$ 0.02 & 0.73 $\pm$ 0.03 \\
DCR & 0.21 $\pm$ 0.02 & \textbf{0.29} $\pm$ 0.02 & 0.71 $\pm$ 0.04 \\
LICEM & \textbf{0.18} $\pm$ 0.01 & 0.39 $\pm$ 0.01 & \textbf{0.58} $\pm$ 0.02 \\
\bottomrule
\end{tabular}
\end{table}

On average, LICEM achieves the lowest deviation from the ground-truth sparsity on the DRUG and MULTIEMO datasets, indicating that its explanations most faithfully reflect the true number of active concepts. DCR also exhibits consistently low deviation across datasets. By contrast, the CBM variants tend to produce higher deviations, particularly on CEBaB and MULTIEMO. These results complement the concept accuracy findings (Tables~\ref{tab:conc_acc_bert} and~\ref{tab:conc_acc}) and underscore the importance of jointly evaluating sparsity and alignment with ground-truth concepts.

\end{appendices}

\end{document}

%% file: new_task_acc.tex
\begin{table*}[t]
\centering
\caption{Task accuracy (\%) of the compared models. We report in \textbf{bold} the best result among the same type of models (e.g., supervised, interpretable ones) considering models equally best if their standard deviations overlap. We use \cmark to indicate models requiring concept supervision (C. S.) or having a task-interpretable predictor (T.I.). We highlight in light gray the models we propose in this work. The Self-Generative and Generative approaches extend the scalability of concept-based models to datasets without concept annotations, where supervised models cannot be applied ($-$).}
\label{tab:task_acc}
{\small
\setlength{\tabcolsep}{2pt}
%\begin{adjustbox}{width=\textwidth}
\resizebox{\textwidth}{!}{
\begin{tabular}{l|lcc|lll|lcccc}
\toprule
Type & Method & C.S. & T.I. & CEBaB & Multiemo-It & Drug & Depression & IMDb & TREC-50 & CLINC-OOS & Banking-77 \\
\midrule
\multirow{3}{*}{\begin{tabular}{@{}l@{}} \textsc{Mixtral}\\ \textsc{(e2e)} \end{tabular}}
& MLP & \xmark & \xmark & \textbf{88.80}\tiny{$\pm$0.75} & 80.01\tiny{$\pm$0.63} & \textbf{63.66}\tiny{$\pm$1.20} & \textbf{97.18}\tiny{$\pm$0.03} & 86.92\tiny{$\pm$0.48} & \textbf{86.33}\tiny{$\pm$0.23} & \textbf{85.14}\tiny{$\pm$0.19} & \textbf{91.81}\tiny{$\pm$0.24} \\
& Zero-Shot & \xmark & \xmark & 86.80\tiny{$\pm$0.31} & 80.06\tiny{$\pm$0.66} & 60.81\tiny{$\pm$0.28} & 73.77\tiny{$\pm$0.23} & 95.05\tiny{$\pm$0.19} & 69.67\tiny{$\pm$0.61} & 82.22\tiny{$\pm$0.43} & 78.34\tiny{$\pm$0.11} \\
& Few-Shot & \xmark & \xmark & 84.79\tiny{$\pm$0.67} & \textbf{84.17}\tiny{$\pm$0.67} & 62.16\tiny{$\pm$0.27} & 76.38\tiny{$\pm$0.08} & \textbf{94.67}\tiny{$\pm$0.00} & 72.13\tiny{$\pm$0.31} & 84.22\tiny{$\pm$0.59} & 78.23\tiny{$\pm$0.11} \\

\midrule
\multirow{7}{*}{\textsc{sup.}} 
& CBM+MLP & \cmark & \xmark & 78.41\tiny{$\pm$9.30} & 45.43\tiny{$\pm$8.20} & 45.42\tiny{$\pm$4.90} & -- & -- & -- & -- & -- \\
& CBM+XG & \cmark & \xmark & 83.01\tiny{$\pm$0.10} & 69.01\tiny{$\pm$0.02} & 55.00\tiny{$\pm$0.13} & -- & -- & -- & -- & -- \\
& CEM & \cmark & \xmark & \textbf{89.60}\tiny{$\pm$0.49} & \textbf{83.33}\tiny{$\pm$0.47} & \textbf{66.81}\tiny{$\pm$0.40} & -- & -- & -- & -- & -- \\
\cmidrule{2-12}
& CBM+LL & \cmark & \cmark & 71.43\tiny{$\pm$9.71} & 42.67\tiny{$\pm$7.01} & 34.60\tiny{$\pm$10.10} & -- & -- & -- & -- & -- \\
& CBM+DT & \cmark & \cmark & 77.20\tiny{$\pm$0.40} & 65.00\tiny{$\pm$0.02} & 47.20\tiny{$\pm$0.40} & -- & -- & -- & -- & -- \\
& DCR & \cmark & \cmark & 88.05\tiny{$\pm$0.53} & 82.01\tiny{$\pm$0.71} & 65.40\tiny{$\pm$0.80} & -- & -- & -- & -- & -- \\
& LICEM & \cmark & \cmark & \textbf{89.89}\tiny{$\pm$0.77} & \textbf{83.47}\tiny{$\pm$0.49} & \textbf{66.80}\tiny{$\pm$0.29} & -- & -- & -- & -- & -- \\

\midrule
\multirow{7}{*}{\textsc{gen.}} 
& CBM+MLP & \xmark & \xmark & 76.19\tiny{$\pm$3.40} & 70.28\tiny{$\pm$0.65} & 42.43\tiny{$\pm$1.10} & 84.72\tiny{$\pm$1.72} & 84.76\tiny{$\pm$0.66} & 49.13\tiny{$\pm$0.23} & 14.30\tiny{$\pm$0.70} & 13.77\tiny{$\pm$0.65} \\
& CBM+XG & \xmark & \xmark & 74.43\tiny{$\pm$0.40} & 69.80\tiny{$\pm$0.09} & 53.85\tiny{$\pm$0.30} & 86.87\tiny{$\pm$0.03} & 69.71\tiny{$\pm$1.06} & 38.47\tiny{$\pm$5.31} & 5.50\tiny{$\pm$7.67} & 3.16\tiny{$\pm$0.97} \\
& CEM & \xmark & \xmark & \textbf{89.97}\tiny{$\pm$0.66} & \textbf{82.41}\tiny{$\pm$0.11} & \textbf{63.80}\tiny{$\pm$0.38} & \textbf{97.06}\tiny{$\pm$0.11} & \textbf{85.97}\tiny{$\pm$0.11} & \textbf{77.67}\tiny{$\pm$1.29} & \textbf{69.66}\tiny{$\pm$0.35} & \textbf{86.31}\tiny{$\pm$0.44} \\
\cmidrule{2-12}
& CBM+LL & \xmark & \cmark & 62.07\tiny{$\pm$0.22} & 68.66\tiny{$\pm$4.20} & 33.14\tiny{$\pm$2.10} & 50.25\tiny{$\pm$0.39} & 82.67\tiny{$\pm$0.66} & 46.33\tiny{$\pm$0.12} & 13.87\tiny{$\pm$0.33} & 10.48\tiny{$\pm$0.07} \\
& CBM+DT & \xmark & \cmark & 78.03\tiny{$\pm$0.23} & 65.32\tiny{$\pm$0.39} & 40.06\tiny{$\pm$1.30} & 83.10\tiny{$\pm$0.13} & 72.00\tiny{$\pm$2.84} & 22.07\tiny{$\pm$18.61} & 1.48\tiny{$\pm$0.26} & 4.08\tiny{$\pm$0.84} \\
& DCR & \xmark & \cmark & 88.97\tiny{$\pm$0.18} & \textbf{80.82}\tiny{$\pm$0.54} & 63.74\tiny{$\pm$1.16} & \textbf{95.35}\tiny{$\pm$0.21} & 84.32\tiny{$\pm$0.29} & 51.20\tiny{$\pm$1.91} & 17.72\tiny{$\pm$0.95} & 39.03\tiny{$\pm$2.52} \\
& \cellcolor{customgray!20}LICEM & \cellcolor{customgray!20}\xmark & \cellcolor{customgray!20}\cmark & \cellcolor{customgray!20}\textbf{90.64}\tiny{$\pm$0.38} & \cellcolor{customgray!20}\textbf{81.85}\tiny{$\pm$0.71} & \cellcolor{customgray!20}\textbf{66.15}\tiny{$\pm$0.44} & \cellcolor{customgray!20}\textbf{95.79}\tiny{$\pm$1.40} & \cellcolor{customgray!20}\textbf{85.46}\tiny{$\pm$0.40} & \cellcolor{customgray!20}\textbf{77.47}\tiny{$\pm$0.70} & \cellcolor{customgray!20}\textbf{72.98}\tiny{$\pm$0.58} & \cellcolor{customgray!20}\textbf{86.66}\tiny{$\pm$0.20} \\
\midrule
\multirow{7}{*}{\begin{tabular}{@{}l@{}}\textsc{self}\\\textsc{gen.}\\\textsc{\scriptsize(ours)}\end{tabular}} 
& \cellcolor{customgray!20}CBM+MLP & \cellcolor{customgray!20}\xmark & \cellcolor{customgray!20}\xmark & \cellcolor{customgray!20}82.71\tiny{$\pm$0.01} & \cellcolor{customgray!20}75.42\tiny{$\pm$4.42} & \cellcolor{customgray!20}47.59\tiny{$\pm$1.37} & \cellcolor{customgray!20}82.31\tiny{$\pm$0.04} & \cellcolor{customgray!20}86.10\tiny{$\pm$0.00} & \cellcolor{customgray!20}46.80\tiny{$\pm$0.00} & \cellcolor{customgray!20}12.31\tiny{$\pm$0.07} & \cellcolor{customgray!20}7.68\tiny{$\pm$0.10} \\
& \cellcolor{customgray!20}CBM+XG & \cellcolor{customgray!20}\xmark & \cellcolor{customgray!20}\xmark & \cellcolor{customgray!20}82.70\tiny{$\pm$1.23} & \cellcolor{customgray!20}79.09\tiny{$\pm\leq$0.01} & \cellcolor{customgray!20}53.28\tiny{$\pm\leq$0.01} & \cellcolor{customgray!20}82.28\tiny{$\pm\leq$0.01} & \cellcolor{customgray!20}72.57\tiny{$\pm$0.00} & \cellcolor{customgray!20}38.40\tiny{$\pm$5.54} & \cellcolor{customgray!20}14.89\tiny{$\pm$0.00} & \cellcolor{customgray!20}3.28\tiny{$\pm$1.41} \\
& \cellcolor{customgray!20}CEM & \cellcolor{customgray!20}\xmark & \cellcolor{customgray!20}\xmark & \cellcolor{customgray!20}\textbf{89.14}\tiny{$\pm$0.38} & \cellcolor{customgray!20}\textbf{84.06}\tiny{$\pm$0.09} & \cellcolor{customgray!20}\textbf{65.20}\tiny{$\pm$0.73} & \cellcolor{customgray!20}\textbf{97.16}\tiny{$\pm$0.08} & \cellcolor{customgray!20}\textbf{88.19}\tiny{$\pm$0.99} & \cellcolor{customgray!20}\textbf{82.47}\tiny{$\pm$1.15} & \cellcolor{customgray!20}\textbf{76.41}\tiny{$\pm$0.40} & \cellcolor{customgray!20}\textbf{88.99}\tiny{$\pm$0.23} \\
\cmidrule{2-12}
& \cellcolor{customgray!20}CBM+LL & \cellcolor{customgray!20}\xmark & \cellcolor{customgray!20}\cmark & \cellcolor{customgray!20}82.71\tiny{$\pm$0.01} & \cellcolor{customgray!20}77.15\tiny{$\pm$0.96} & \cellcolor{customgray!20}47.35\tiny{$\pm$0.29} & \cellcolor{customgray!20}82.12\tiny{$\pm$0.15} & \cellcolor{customgray!20}84.95\tiny{$\pm$1.98} & \cellcolor{customgray!20}45.67\tiny{$\pm$0.12} & \cellcolor{customgray!20}12.40\tiny{$\pm$0.31} & \cellcolor{customgray!20}7.88\tiny{$\pm$0.10} \\
& \cellcolor{customgray!20}CBM+DT & \cellcolor{customgray!20}\xmark & \cellcolor{customgray!20}\cmark & \cellcolor{customgray!20}83.95\tiny{$\pm\leq$0.01} & \cellcolor{customgray!20}78.44\tiny{$\pm\leq$0.01} & \cellcolor{customgray!20}53.28\tiny{$\pm\leq$0.01} & \cellcolor{customgray!20}82.28\tiny{$\pm\leq$0.01} & \cellcolor{customgray!20}86.10\tiny{$\pm$0.00} & \cellcolor{customgray!20}46.80\tiny{$\pm$0.00} & \cellcolor{customgray!20}12.53\tiny{$\pm$0.00} & \cellcolor{customgray!20}8.34\tiny{$\pm$0.00} \\
& \cellcolor{customgray!20}DCR & \cellcolor{customgray!20}\xmark & \cellcolor{customgray!20}\cmark & \cellcolor{customgray!20}87.72\tiny{$\pm$0.66} & \cellcolor{customgray!20}83.47\tiny{$\pm$0.43} & \cellcolor{customgray!20}63.29\tiny{$\pm$0.36} & \cellcolor{customgray!20}97.11\tiny{$\pm$0.03} & \cellcolor{customgray!20}\textbf{88.83}\tiny{$\pm$0.40} & \cellcolor{customgray!20}\textbf{82.20}\tiny{$\pm$0.69} & \cellcolor{customgray!20}74.90\tiny{$\pm$0.63} & \cellcolor{customgray!20}88.55\tiny{$\pm$0.37} \\
& \cellcolor{customgray!20}LICEM & \cellcolor{customgray!20}\xmark & \cellcolor{customgray!20}\cmark & \cellcolor{customgray!20}\textbf{89.56}\tiny{$\pm$0.29} & \cellcolor{customgray!20}\textbf{84.49}\tiny{$\pm$0.25} & \cellcolor{customgray!20}\textbf{65.89}\tiny{$\pm$0.39} & \cellcolor{customgray!20}\textbf{97.23}\tiny{$\pm$0.02} & \cellcolor{customgray!20}\textbf{88.44}\tiny{$\pm$0.22} & \cellcolor{customgray!20}80.00\tiny{$\pm$1.06} & \cellcolor{customgray!20}\textbf{76.90}\tiny{$\pm$0.40} & \cellcolor{customgray!20}\textbf{89.29}\tiny{$\pm$0.25} \\
\bottomrule
\end{tabular}
}
}
\end{table*}

%% file: conc_acc.tex
% \begin{table}[t]
\begin{table}[t]%{R}{9.cm}
\centering
\caption{This table presents the performance in terms of concept prediction of the models that utilize \hbox{Mixtral 8x7B} as backbone. Concept prediction (\%) of the compared models for datasets equipped with concept annotations is measured using the macro-averaged F1 score. We report in \textbf{bold} the best result among the same type of models (e.g., supervised, interpretable ones) considering models equally best if their standard deviations overlap. Self-supervised methods are reported with the same concept accuracy with zero standard deviation, since the concept predictions are provided by an LLM with temperature set to zero. The methods using the self-generative have the same macro-averaged F1 score, therefore we use $-$ to represent all methods.}
\label{tab:conc_acc}
% \resizebox{9cm}{!}{
\begin{tabular}{l|l|cccc}
\toprule
Type & Method        & CEBaB                               & Multiemo-It        &               Drug    \\
\midrule
\textsc{e2e}
& MLP            & \textbf{75.92}\tiny{$\pm$ 0.77}                      & \textbf{74.25}\tiny{$\pm$ 1.02}        &  \textbf{78.50}\tiny{ $\pm$0.23}  \\
\midrule
\multirow{7}{*}{\textsc{sup.}} %{\rotatebox[origin=c]{90}{\textsc{supervised}}}
& CBM+MLP       & 65.17\tiny{$\pm$ 2.35}                      & 61.75\tiny{$\pm$ 1.02}         & 65.33\tiny{$\pm$2.46} \\ 
& CEM           & \textbf{78.83}\tiny{$\pm$ 0.85}                      & \textbf{77.12}\tiny{$\pm$ 1.38}        & \textbf{80.79}\tiny{$\pm$0.47} \\ 
& CBM+XG           & 75.92\tiny{$\pm$ 0.77}                 & 74.25\tiny{$\pm$ 1.02}         & 78.50\tiny{$\pm$ 0.23} \\ 

\cmidrule{2-5}
& CBM+LL        & 64.25\tiny{$\pm$ 2.56}                      & 59.12\tiny{$\pm$ 2.13}       & 64.83\tiny{$\pm$ 1.20} \\ 
& CBM+DT        & \textbf{75.92}\tiny{$\pm$ 0.77}                   & 74.25\tiny{$\pm$ 1.02}       & 78.50\tiny{$\pm$0.23} \\ 
& DCR       & \textbf{78.45}\tiny{$\pm$ 1.92}                     & \textbf{75.67}\tiny{$\pm$ 1.43}              & 79.96\tiny{$\pm$ 0.43}       \\ 
& \cellcolor{customgray!20}LICEM (ours)        & \cellcolor{customgray!20}75.45\tiny{ $\pm$ 0.93}        & \cellcolor{customgray!20}\textbf{76.36}\tiny{ $\pm$ 0.39}        & \cellcolor{customgray!20}\textbf{80.83}\tiny{ $\pm$ 0.36} \\ 
\midrule
\multirow{7}{*}{\textsc{gen.}}%\rotatebox[origin=c]{90}{\textsc{generative}}} 
& CBM+MLP       & 71.87\tiny{ $\pm$ 0.14}                      & 52.60\tiny{ $\pm$ 14.32}          & 55.68\tiny{ $\pm$ 19.84} \\ 
& CEM           & \textbf{74.70}\tiny{ $\pm$ 0.98}                      & \textbf{63.61}\tiny{ $\pm$ 0.44}  & \textbf{79.45}\tiny{ $\pm$ 0.41} \\
& CBM+XG           & \textbf{75.02}\tiny{ $\pm$ 0.57}          & 61.69\tiny{ $\pm$ 0.44}            & \textbf{79.15}\tiny{ $\pm$ 0.30} \\ 
\cmidrule{2-5}
& CBM+LL        & 72.15\tiny{ $\pm$ 0.59}                    & \textbf{63.72}\tiny{ $\pm$ 0.84}         & 66.72\tiny{ $\pm$ 19.48}  \\
& CBM+DT           & \textbf{75.02}\tiny{ $\pm$ 0.57}                      & 61.69\tiny{ $\pm$ 0.44}        & \textbf{79.04}\tiny{ $\pm$ 0.30} \\ 
& DCR       & \textbf{75.62}\tiny{ $\pm$ 2.59}                    & 62.79\tiny{ $\pm$ 0.44}         & \textbf{79.04}\tiny{ $\pm$ 0.33} \\ 
& \cellcolor{customgray!20}LICEM (ours)        & \cellcolor{customgray!20}\textbf{74.44}\tiny{ $\pm$ 0.25}                    & \cellcolor{customgray!20}\textbf{63.75}\tiny{ $\pm$ 0.36}         & \cellcolor{customgray!20}\textbf{79.05}\tiny{ $\pm$ 0.58}\\ 

\midrule
\multirow{1}{*}{\begin{tabular}{@{}l@{}}\textsc{self gen.}\end{tabular}} 
& \cellcolor{customgray!20}--   & \cellcolor{customgray!20}\textbf{84.08}\tiny{ $\pm$ 0.00}                    & \cellcolor{customgray!20}\textbf{64.27}\tiny{ $\pm$0.00}          & \cellcolor{customgray!20}\textbf{83.00}\tiny{ $\pm$0.00} \\ 
\bottomrule
\end{tabular}
% }
\end{table}